\documentclass{article}


\usepackage{graphicx}
\usepackage[margin=0.7in]{geometry}
\usepackage{amsmath,amssymb,amsfonts}
\usepackage{algorithmic}
\usepackage{textcomp}
\usepackage{verbatim}
\usepackage{booktabs}
\usepackage{multirow}
\usepackage{tabularray}
\usepackage{tabularx}
\usepackage{subcaption}
\usepackage{tabu}
\usepackage{siunitx}
\usepackage[colorlinks=true, linkcolor=blue, urlcolor=blue]{hyperref}
\usepackage{makecell}
\usepackage[table]{xcolor}
\usepackage{cite}

\title{RealSynCol: a high-fidelity synthetic colon dataset for 3D reconstruction applications}
\date{} %

\author{
    Chiara Lena$^{1 *}$, Davide Milesi$^{1}$, Alessandro Casella$^{1}$, \\ Luca Carlini$^{1}$, Joseph C. Norton$^{2}$, James Martin$^{2}$,\\ Bruno Scaglioni$^{2}$, Keith L. Obstein$^{3, 4}$, Roberto De Sire$^{5,6}$,\\ Marco Spadaccini$^{5,6}$, Cesare Hassan$^{5,6}$, Pietro Valdastri$^{2}$,\\ Elena De Momi$^{1}$ 
}

\date{
    \small
    $^{1}$Department of Electronics, Information, and Bioengineering, Politecnico di Milano, Milan, Italy\\
    $^{2}$School of Electronic and Electrical Engineering, University of Leeds, Leeds, UK\\
    $^{3}$Division of Gastroenterology, Vanderbilt University Medical Center, Nashville, TN, USA\\
    $^{4}$Department of Mechanical Engineering, Vanderbilt University, Nashville, TN, USA\\
    $^{5}$Endoscopy Unit, Humanitas Clinical and Research Center IRCCS, Rozzano, Italy\\
    $^{6}$Department of Biomedical Sciences, Humanitas University, Pieve Emanuele, Italy\\
    $^{*}$Corresponding author: \texttt{chiara.lena@polimi.it}
}

\begin{document}

\maketitle

\begin{abstract}
Deep learning has the potential to improve colonoscopy by enabling 3D reconstruction of the colon, providing a comprehensive view of mucosal surfaces and lesions, and facilitating the identification of unexplored areas. However, the development of robust methods is limited by the scarcity of large-scale ground truth data. 
We propose RealSynCol, a highly realistic synthetic dataset designed to replicate the endoscopic environment. Colon geometries extracted from 10 CT scans were imported into a virtual environment that closely mimics intraoperative conditions and rendered with realistic vascular textures. The resulting dataset comprises 28\,130 frames, paired with ground truth depth maps, optical flow, 3D meshes, and camera trajectories.
A benchmark study was conducted to evaluate the available synthetic colon datasets for the tasks of depth and pose estimation. Results demonstrate that the high realism and variability of RealSynCol significantly enhance generalization performance on clinical images, proving it to be a powerful tool for developing deep learning algorithms to support endoscopic diagnosis.

\end{abstract}

\section{Background \& Summary}
Colorectal cancer (CRC) remains the second leading cause of cancer-related mortality worldwide. However, recent epidemiological studies have highlighted a sustained decline in its incidence and mortality~\cite{Siegel2023}. This trend is attributed to preventive screening procedures, which enable the detection and removal of lesions in precancerous or early-stage phases~\cite{LADABAUM2020}, for which the five-year survival rate exceeds 90\%~\cite{chen2017}. Colonoscopy, performed with a flexible endoscope, remains the gold standard for CRC screening, although magnetic endoscopes are emerging as promising alternatives~\cite{Martin2020}. 

Nonetheless, diagnostic outcome highly depends on the endoscopist’s expertise, with an estimated 25\% of neoplastic lesions still undetected~\cite{nishihara2013}. The primary cause is the incomplete examination of the colonic surface. The peculiar colon topology, characterized by haustral folds and loops in the tissues, combined with the limited maneuverability of the endoscope, can hamper the visualization of the mucosa, leading to regions not adequately inspected~\cite{ma2021}. A study using virtual colonoscopy~\cite{hong2007} demonstrated that approximately 23\% of the mucosa may be missed during standard procedures. 

Real-time three-dimensional (3D) reconstruction of the colon can assist in identifying blind spots, thereby reducing the risk of missed lesions and alerting physicians to re-examine insufficiently inspected regions. 
Monocular 3D reconstruction has been widely investigated, with recent methods exploiting deep learning algorithms~\cite{Fu2021}. These can be integrated into clinical practice without additional burden on endoscopists and may increase diagnostic yield by reducing reliance on operator-dependent visualization, while also being cost-effective. 

The development of accurate 3D reconstruction algorithms is limited by the need for large volumes of annotated data, the acquisition of which is often infeasible in clinical practice. To address this challenge, synthetic datasets have emerged as promising alternatives. Nonetheless, differences in image characteristics between synthetic and real-world data pose significant challenges for domain transfer.

\subsection{Synthetic colon datasets}\label{sec:synthetic_datasets}
Synthetic colon datasets are typically generated with a setup closely resembling real endoscopic conditions, characterized by a limited field of view, restricted movement range, and a sole light source coupled with a monocular camera mounted on the tip of the device. 
Accurately reproducing the characteristics of intra-operative images remains challenging due to the high variability in texture, lesions, and tissue colors. Moreover, the introduction of water into the lumen during the procedure often leads to specular reflections.

In~\cite{Mahmood2018}, the first synthetic dataset for colonoscopy is introduced, comprising colon models rendered within a 3D environment, from which over 100\,000 RGB frames and the corresponding depth maps are extracted.

The dataset in~\cite{Rau2019} aims at reducing the gap between real and synthetic images by increasing variability. The surface mesh is obtained from a Computed Tomography (CT) scan of a real colon, through manual segmentation and meshing. The model is imported into Unity\textregistered{} (Unity Technologies, San Francisco, CA \cite{unity}), and a virtual camera with two light sources is generated. The dataset includes 16\,016 images and the corresponding ground truth depth maps, with different configurations of light, material, and camera movements.

However, the lack of a realistic texture, the limited shape variability (just one colon model is used) and camera trajectories, and the small image size (256x256 pixels, compared to HD or Full-HD colonoscopy images) remain open challenges.

A similar data generation process is described for EndoSLAM~\cite{EndoSLAM}. CT scans are rendered in a simulation environment and processed to reduce artifacts~\cite{Incetan2020}. By navigating the model, a colon sequence with frames paired with depth and poses is obtained. 

The challenge of realistic texture is addressed by Sinth-colon, which includes 20\,000 images with corresponding depth and 3D position~\cite{Synth-colon}. A CycleGAN is trained using the public dataset Kvasir~\cite{kvasir} to generate texture images. The shape is sculpted in 3D rather than derived from CT scans, resulting in reduced anatomical realism. 

Endomapper~\cite{Endomapper} exploits the method for virtual environment creation described in~\cite{Incetan2020} to generate sequences with varying deformation parameters. The synthetic dataset includes ground truth depth maps, trajectories, camera calibration files, camera parameters, and units of deformation. The released version consists of 6 sequences depicting the same colon tract and only differing in deformation parameters, thus, data variability is low. 

In C3VD~\cite{C3VD}, colonoscopy references are used to digitally sculpt a 3D model and print the relative mold. Image acquisition is performed with a real colonoscope (Olympus\textregistered{}  CF-HQ190L) mounted on a UR-3 (Universal Robotics\textregistered{}) robotic arm. A CycleGAN \cite{Zhou2017} is trained to predict depth maps from the acquired RGB frames, and raw poses are extracted from the robot. Finally an optimization procedure is performed: 2D frames are aligned with the 3D virtual model to compute the transformation; for each camera pose, depth maps are rendered in the virtual model and compared to the predicted depth frame to increase the accuracy of depth and transformation predictions. 22 sequences and 10\,015 frames were recorded, each with depth map, occlusion map, optical flow, and camera pose. Every video sequence is paired with the ground truth surface and the coverage map. However, the camera movements are significantly shorter and slower compared to real colonoscopy, resulting in an unrealistic representation of motion.  

The most recent dataset is SimCol3D~\cite{simcol}. Images are created using CT scans from three patients, including the colon model used to generate the dataset in~\cite{Rau2019}. The scans are processed in Unity\textregistered{}, and 33 trajectories are acquired. 
Depth maps, camera poses, and camera settings are also provided. Despite the addition of small random perturbations, all trajectories acquired within the same model appear relatively similar, exhibiting smooth motion along the centerline. Moreover, all sequences display movements exclusively in the forward direction. 

A schema of the described monocular datasets is depicted in Table~\ref{tab:dataset_analysis}.

\begin{table*}[!ht]
\caption{Overview of the monocular synthetic colon datasets described in Sec.~\ref{sec:synthetic_datasets}. RealSynCol is included in the last row (highlighted in light blue) for comparison. The column "\# Colon Shapes" reports the number of distinct patients or colon shapes used for dataset generation. The asterisk (*) indicates that in C3VD only one colon shape was used, but four different physical phantoms were created, each with distinct textures and colors.}
\label{tab:dataset_analysis}
\resizebox{\textwidth}{!}{
\begin{tabular}{cccccccc}
\textbf{Dataset} & \textbf{Frames} & \textbf{\# Colon Shapes} & \textbf{Resolution} & \textbf{Depth map} & \textbf{Optical Flow} & \textbf{Camera Pose} & \textbf{Public} \\ \hline
Mahmood \textit{et al.}~\cite{Mahmood2018} & \textgreater{100\,000} & 1 & 720x576 & yes & no & no & no \\ \hline
Rau \textit{et al.}~\cite{Rau2019} & 16\,016  & 1 & 256x256 & yes & no & no & yes \\ \hline
EndoSLAM~\cite{EndoSLAM} &  21\,887  & 1  & 320x320 & yes & no & yes & yes \\ \hline
Synth-colon~\cite{Synth-colon}  & 20\,000 & 1 & 500x500 & yes & no & no & yes \\ \hline
Endomapper~\cite{Endomapper} & 2\,017 & 1 & 960x720 & yes & no & yes & yes \\ \hline
C3VD~\cite{C3VD} & 10\,015 & 1* & 1350x1080 & yes & yes & yes & yes \\ \hline
SimCol3D~\cite{simcol} & 23\,421 & 3 & 475x475 & yes & no & yes & yes \\ \hline
\rowcolor{cyan!20} 
RealSynCol & 28\,130 & 10 & 1024x1024 & yes & yes & yes & yes \\ \hline
\end{tabular}
}
\end{table*}

\subsection{3D Reconstruction in endoscopy}
3D reconstruction in endoscopy remains an ongoing challenge, driving active research in the field. Due to its inherent complexity, the problem is commonly divided into depth estimation and camera pose estimation~\cite{ma2019real, lena2024}. Both require large-scale, annotated datasets. However, the characteristics of the endoscopic environment, including limited field of view, specular reflections, frequent occlusions, and the absence of dedicated sensors on commercial endoscopes, significantly hinder the direct acquisition of depth and pose information. Current research heavily relies on deep learning approaches, which further amplifies the demand for high-quality annotated data.
To address the scarcity of ground truth data, researchers have explored alternative approaches. 

One direction involves the use of synthetic datasets to replicate the missing information~\cite{Rau2019, EndoSLAM, Endomapper, C3VD, simcol}. In~\cite{Mahmood2018}, a convolutional neural network is trained on a synthetic dataset to estimate depth and colon topography. In~\cite{Rau2019}, depth estimation is improved by integrating synthetic images with unlabelled frames to facilitate adaptation to real-world scenarios. Similarly,~\cite{Itoh2021} employs an ideal Lambertian model to generate an RGB-D dataset that combines real and virtual images, enabling the network to learn domain translation.

Another line of research explores Structure from Motion (SfM) and Simultaneous Localization and Mapping (SLAM) approaches, often combined with neural networks. The SLAM approach in~\cite{chen2019} merges RGB images with depth maps predicted with an adversarially-trained neural network to obtain a dense reconstruction. In~\cite{ma2019real} and \cite{ma2021}, a deep learning-based SLAM pipeline is proposed to reconstruct 3D portions and estimate missed areas, leveraging a recurrent neural network for depth and pose prediction. During training, sparse depth maps from SfM serve as ground truth, whereas the estimated poses are refined using a direct SLAM approach.
Complementary to the aforementioned methods, several studies starting from Monodepth2~\cite{monodepth2}, have investigated self-supervision for jointly learning depth and pose representations from unlabeled images, using a minimum reprojection loss as supervisory signal. Building on Monodepth2, multiple approaches have been proposed for natural scenes, such as Mono-ViT~\cite{zhao2022} and Lite-Mono~\cite{zhang2023}. EndoSfM-Learner~\cite{EndoSLAM} introduces a brightness-aware photometric loss to compensate for the illumination changes typical of endoscopic environments. The challenge of brightness inconsistency is also addressed in~\cite{shao2021} and~\cite{shao2022}, where an appearance flow module accounts and compensates for variations in brightness patterns. Recently, different approaches have exploited the self-supervision paradigm in conjunction with foundation models to increase the reliability of depth predictions~\cite{cui2024, zeinoddin2024}. 

\subsection{Potential research gaps and motivation}
Although several synthetic datasets have been introduced in recent literature, a substantial gap remains between synthetic images and the visual characteristics of real clinical scenarios. This domain discrepancy continues to pose a major challenge for learning-based methods, especially in pose estimation and related 3D reconstruction tasks, where limited anatomical realism and low variability in texture or motion prevent synthetic-trained models from generalizing effectively to real endoscopic data. Existing datasets often rely on a small number of different colon models and exhibit texture and motion patterns that diverge from those observed in clinical practice.

To mitigate this domain gap, we introduce RealSynCol, a high-fidelity synthetic dataset that accurately replicates the anatomy, endoscope motion patterns, and visual appearance of the colon environment. The dataset comprises 28\,130 frames organized into 20 sequences derived from 10 anatomically realistic colon models. Each frame includes high-quality depth and optical-flow annotations, while each sequence provides camera intrinsic parameters, ground-truth trajectories, and the corresponding 3D mesh. Clinical measurements, samples, and imaging data were used as references during dataset generation to ensure a level of realism that closely resembles real clinical conditions.

In addition, this study benchmarks the main synthetic datasets currently available for colonoscopy and presents an ablation analysis to evaluate how different components and characteristics of synthetic imagery influence the learning process.

\section{Methods}\label{method}
\begin{figure*}[t]
\includegraphics[width=0.9\textwidth]{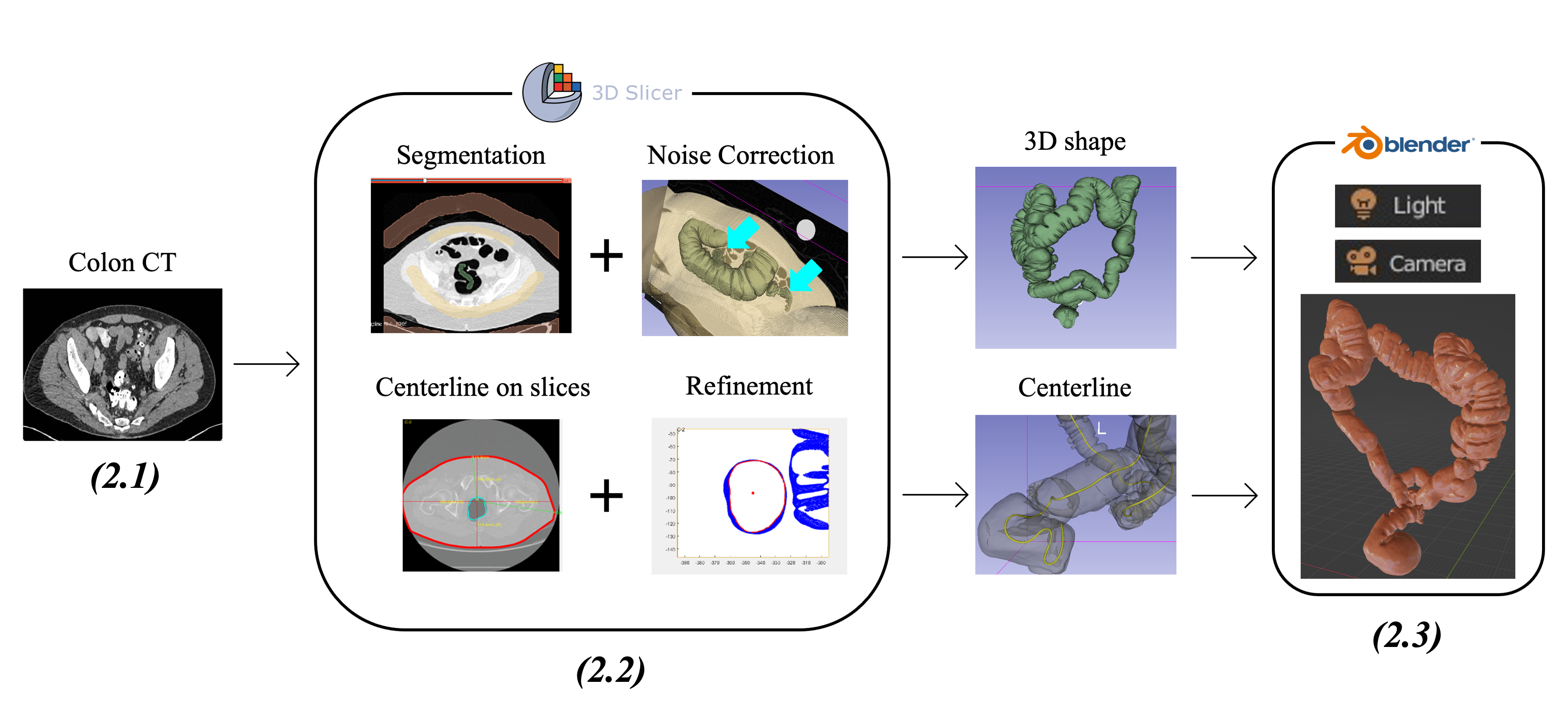}
\centering
\caption{Schema of the dataset generation process. CT scans in DICOM files are imported into 3D Slicer~\cite{FEDOROV2012}, segmented, and any noise left is corrected. The centerline is identified and refined in each. The corrected 3D model and the centerline are imported into Blender, where texture, light, and a virtual camera are added. This environment is then used to extract RGB frames, ground truth depth, optical flow, camera parameters, 3D mesh, and trajectory.}
\label{fig:generation}
\end{figure*}

The dataset creation method is outlined in Fig.~\ref{fig:generation}. First, CT images and the processing techniques are described in Sec.~\ref{sec:data_source} and in Sec.~\ref{sec:processing_CT}. Then, the creation of a proper virtual environment is outlined in Sec.~\ref{sec:virtual_env}, and the ground truth generation process is presented in Sec.~\ref{sec:groundtruth_gen}.

\subsection{Colon models}\label{sec:data_source}
The Cancer Imaging Archive CT Colonography (ACRIN 6664) repository was used for this work~\cite{clark2013}, in compliance with the terms of use of the original data sources. This contains CT scans in DICOM format of 825 adults in both prone and supine positions. Following the work in~\cite{norton2024}, a subset of 75 patients was considered, ensuring accepted images met the following criteria: (1) no large artifacts or errors in the CT images; (2) complete patient data; (3) minimal water/feces volume in the bowel lumen (less than one-third of the cross-section). Among these, 10 CT scans were selected to generate the RealSynCol dataset. The selection process was designed to ensure that the resulting subset is representative of the original repository while maintaining substantial variability in anatomical characteristics such as shape, length, diameter, volume, and tortuosity across the final images. Of the 10 CT scans, five were obtained from male patients and five from female patients. Five scans were acquired in the prone position, while the remaining five were acquired in the supine position. The mean patient age was 56 years ($\pm$ 7.04), and the corresponding average colon measurements are summarized in Table~\ref{tab:colon-parameters}.

\begin{table}[!ht]
\centering
\caption{Summary of colon parameters across the 10 selected patients using the measurement protocol established in~\cite{norton2024}.}
\label{tab:colon-parameters}

{%
\sisetup{
  separate-uncertainty = true,
  uncertainty-separator = {\,\pm\,},
}

\resizebox{0.45\columnwidth}{!}{%
\begin{tabular}{l c}
\hline
\textbf{Colon parameter} & \textbf{Average Measurement} \\ \hline
Diameter (cm)           & \num{3.53(32)}        \\
Total length (cm)       & \num{170.07(1633)}    \\
Total volume (cm$^3$)   & \num{2007.29(50637)}  \\
Number of flexures      & \num{16.10(360)}      \\
\hline
\end{tabular}%
}%
}%

\end{table}

\subsection{CT images processing}\label{sec:processing_CT}
CT scans were processed to extract relevant data using 3D Slicer~\cite{FEDOROV2012}. At first, images are segmented in a semi-automatic manner to identify the colon lumen and distinguish it from the patient's abdomen and the surrounding air.  
A subsequent correction step is performed to refine the shape and avoid discontinuities or inaccuracies in the anatomical structures. In each slice, the center of the large bowel is labeled with a fiducial marker. 
The centerline is used to determine the camera's trajectory within the colon. To avoid potential artifacts caused by residual noise, the sequence of points is manually refined to ensure that the centerline is accurately aligned with the center of the lumen. Finally, the refined 3D shape of the large bowel lumen is represented by a dense point cloud. 

\subsection{Virtual models generation}\label{sec:virtual_env}

\subsubsection{Trajectory generation}
\begin{figure} 
\centering 
\includegraphics[width=0.55\columnwidth]{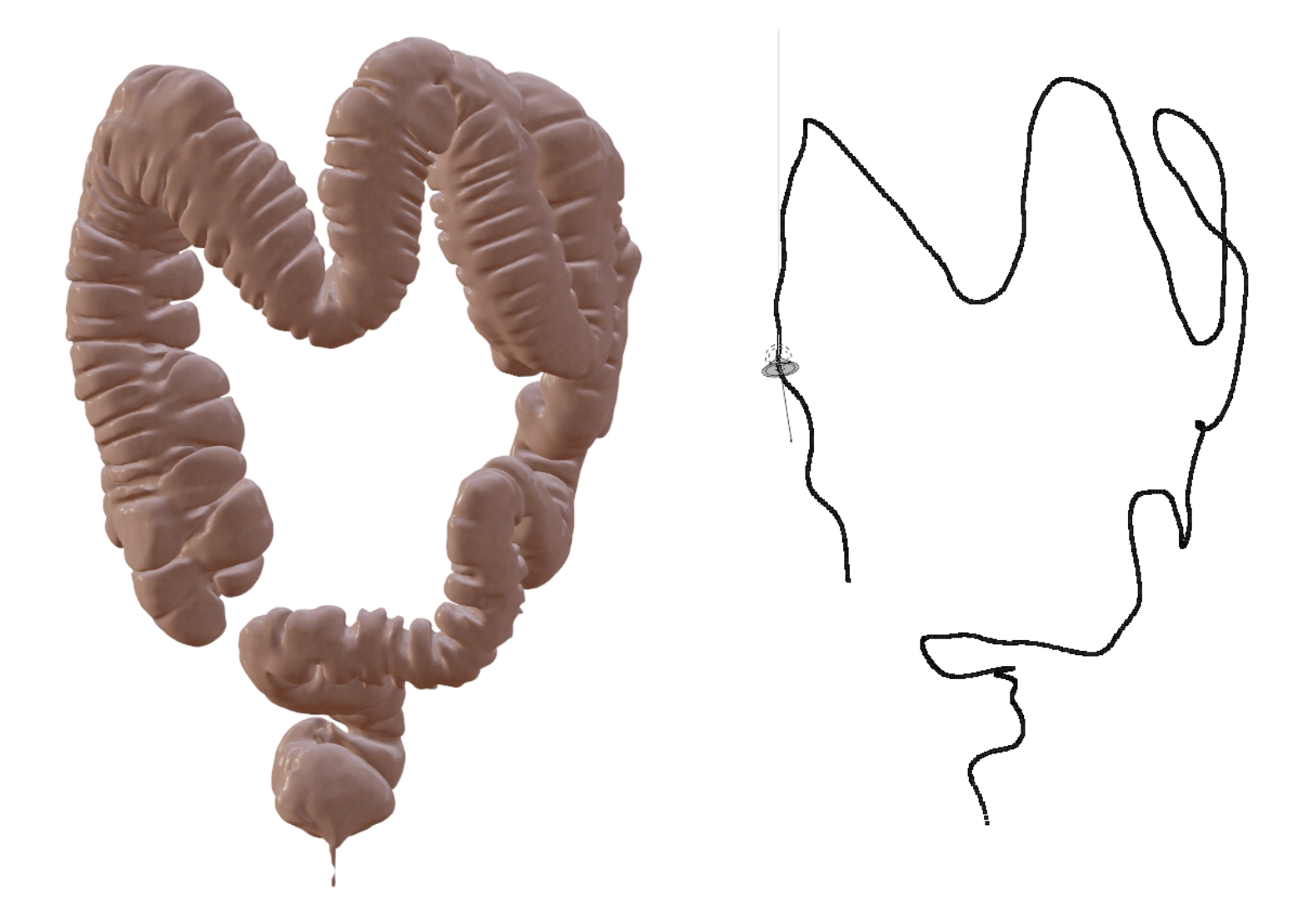}
\caption{Example of a 3D colon model and the relative centerline in Blender}
\label{fig:Blender}
\end{figure}

The 3D colon shapes and the relative centerlines were then imported into Blender\textregistered{} \cite{blender} (Blender version 3.6.9, Blender Foundation, Amsterdam, The Netherlands). A light source and a virtual pinhole camera were designed to reproduce the illumination conditions and imaging of a colonoscope during real procedures, yielding a realistic appearance consistent with clinical colonoscopic views. An example of a 3D colon shape and the relative centerline is depicted in Fig. \ref{fig:Blender}.

The centerline is typically adopted as the trajectory along which the camera is moved within the virtual environment, thus reproducing the navigation of the endoscope and the acquisition of RGB frames and additional data. However, moving the camera along the centerline obtained with the method described in Sec.~\ref{sec:data_source} and pointing in the direction of its trajectory's tangent, would result in an excessively smooth motion, deviating significantly from the real examination scenario. 

Therefore, to generate a more realistic motion profile, we first quantified the amplitude and variability of endoscopic maneuvers in a controlled, clinically inspired setting: an experienced clinician performed endoscope navigation within a deformable silicone phantom, specifically designed to reproduce the anatomical and mechanical properties of the colon, and the measured motion patterns were then used as a reference for realistic trajectory simulation. Following the protocol described in~\cite{Finocchiaro2023}, the phantom was fabricated using Ecoflex™ 00-50 silicone rubber to mimic the compliance and elasticity of colonic tissue. The process involved mold preparation with a separating agent, pigment mixing for realistic coloration, vacuum degassing to eliminate air bubbles, and controlled casting with an 18-minute pot life. After a three-hour curing phase, the mold was opened and cleaned to obtain the final phantom. 
All acquisitions were performed using an Olympus\textregistered{} Exera II CLV-180 endoscopic platform~\cite{olympus}, including an Olympus\textregistered{} PCF-H180AL pediatric colonoscope equipped with a high definition camera in the tip, a xenon-based light source enabling both white-light and narrow-band imaging illumination, and an Olympus\textregistered{} CV-180 video processor with high definition output. An electromagnetic sensor was mounted at the distal tip of the endoscope to continuously record the instrument’s trajectory, capturing both position and orientation at each acquired frame. For pose and orientation acquisition, the NDI\textregistered{} Aurora electromagnetic tracking system was employed~\cite{ndi_aurora_sensors}. The system outputs six degrees of freedom camera poses consisting of 3D position $(x,y,z)$ and orientation $(\phi, \theta, \psi)$ at video rate.

During the experiment, the clinician was instructed to navigate the phantom in a manner closely resembling the withdrawal phase of a standard colonoscopy. The resulting trajectory data were then analyzed to characterize the range and variability of clinically representative endoscopic movements. Three sequences for a total of 3\,213 frames were acquired, yielding a median relative displacement of 0.81 mm (IQR: 1.30 mm) across the three axes. Based on this observed range of motion and its variability, we introduced multiple frame steps during the generation of RealSynCol. Each model was visualized in Blender\textregistered{} over a fixed number of 8\,000 frames, with the camera moving at a constant speed. However, the images corresponding to the initial and final segments of each colon shape were discarded to mitigate CT-related artifacts typically observed in these regions, ensuring that only meaningful frames from the cecum to the rectum were retained. The remaining frames correspond to the points along the camera trajectory, from which a sampling step was applied to reduce redundancy due to highly similar consecutive frames.

To increase the diversity of camera motion, in accordance with the previously discussed findings, a frame step of 2 was applied to two colon models, a step of 5 to five models, and a step of 10 to the remaining three. This configuration yielded translation ranges that closely match those typically observed in clinical endoscopic procedures.

In addition to trajectory subsampling, we introduced controlled noise along the camera trajectory to induce slight inter-frame rotations, making the relative motion between consecutive frames appear less smooth and therefore more realistic with respect to real clinical acquisitions. An invisible Blender “empty” object was created and placed in front of the camera, aligned with the centerline, and kept at a fixed relative distance. The object’s position served as an anchor for the camera orientation, allowing the addition of noise while preserving the underlying trajectory, which coincides with the colon centerline. Experimentally, a random value drawn from a uniform distribution within the range $[-0.5, +0.5]$ mm was added independently to each of the $x$, $y$, and $z$ coordinates of the invisible object along its entire trajectory. 
Since colon phantoms are usually shorter and exhibit fewer tissue bendings with less pronounced haustral folds, the rotational movements required to properly inspect the mucosal surface are generally smaller than those observed in real colonoscopic procedures. Therefore, following the feedback and recommendations of experienced clinicians, we tuned the noise parameters affecting the camera orientation to generate slightly larger rotations than those recorded on the physical phantom. This approach enabled the creation of a dataset that encompasses both subtle and broader rotational motions, representative of realistic maneuvers required to visualize the mucosa in regions with tight bends or lesions located on the haustral folds. As a result, the RealSynCol dataset exhibits a wider range of motion patterns compared to existing synthetic datasets in the literature.

\subsubsection{Texture modeling}
Given the inherent variability of vascular appearance across patients, ranging from well-defined to barely perceptible patterns, and the influence of factors such as tissue deformation, illumination, and fluid artifacts, achieving an exact replication of the colonic texture is not feasible. Instead, we generated realistic yet generalizable textures derived from segmented vascular patterns and real color statistics from clinical images. This approach provides biologically plausible visual cues while maintaining flexibility for algorithmic generalization across diverse anatomical and imaging conditions.

For the texture generation, different vessel patterns were manually segmented from the public SUN Database~\cite{SUN}. Then, a larger non-uniform vascular pattern was obtained using a texture synthesis technique called quilting. Patches extracted from the original pattern were stitched together using the approach of minimum error boundary cut, which minimizes the cost path through the error surface, ensuring the cut between two adjacent patches is in correspondence with pixels with the minimum overlap error~\cite{Efros2001}. 

From images depicting clear mucosa in the public SUN Database~\cite{SUN} (i.e. without lesions, debris, or blood), colors corresponding to vessels and tissues were sampled. Multiple vessel layers were then generated by applying flips, rotations, and Gaussian blurring (using a kernel size of 3 for the first two layers and 1 for the third) to smooth the edges. The final texture was obtained by overlaying the three vessel layers onto a non-uniform background created using the realistic color palette, followed by the application of a Gaussian blur with a kernel size of 5.

\subsection{Ground truth data generation}\label{sec:groundtruth_gen}
\begin{figure}[!htbp]
\centering
\includegraphics[width=0.6\columnwidth]{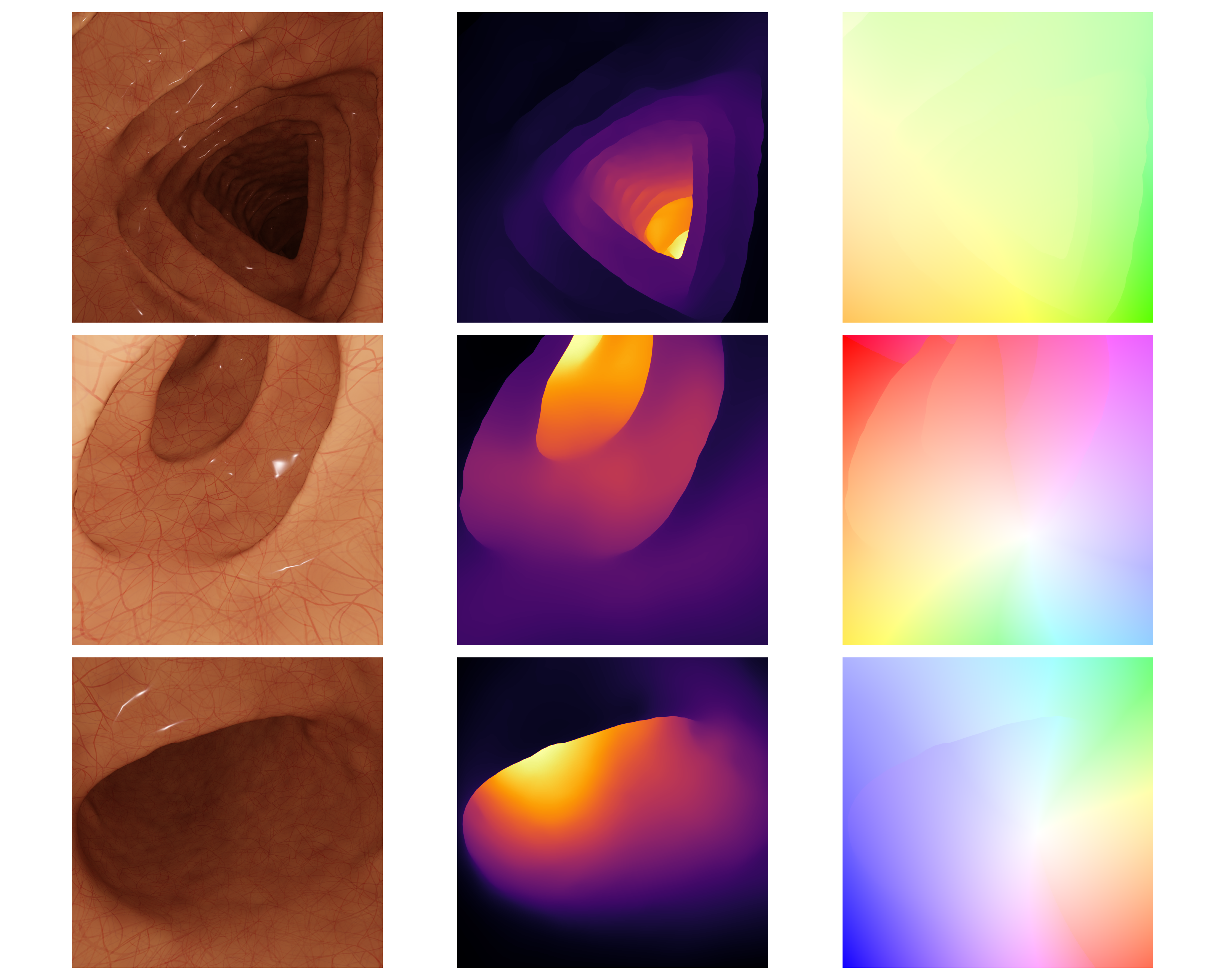}
\caption{Examples of generated images with the relative ground truths. On the left, RGB frames, in the center the depth maps, on the right the optical flow in Middlebury color coding \cite{baker2011}.}
\label{fig:example}
\end{figure}

For each frame, the ground truth depth maps, optical flows, camera intrinsics, and trajectory were computed. The RGB images, depth maps, and optical flows are provided at a resolution of 1024$\times$1024 pixels. An example of the generated images is depicted in Fig.~\ref{fig:example}.

Following the procedure described in Sec.~\ref{sec:virtual_env}, three different sampling steps were employed during data generation, and random noise was added to the camera orientation to better emulate the motion characteristics observed in real endoscopic procedures. A sampling step of 2 was used for two colon models, a step of 5 for five models, and a step of 10 for the remaining three, ensuring a diverse range of motion trajectories consistent with realistic endoscopic movement patterns.

Two trajectories were acquired for each colon shape. The first, in the backward direction, mimics the movement of the endoscope in normal clinical practice, with the camera moving from the cecum to the anus and oriented towards the cecum. The second trajectory, in the forward direction, is acquired with the camera moving from the anus to the cecum and oriented towards the cecum. This allows for increased variability in camera movements.

\section{Data Records}\label{dataset_struct}
The dataset contains 10 different colon models. For each, 2 trajectories are recorded. This leads to 20 sequences and 28\,130 frames. The dataset structure is detailed in Table \ref{tab:structure}.
For each sequence, the dataset provides:
\begin{itemize}
  \item Rendered frames, \textit{Frame/Frame\_XXXX.png}, as 8-bit RGB images.
  \item Depth maps, \textit{Depth/Depth\_XXXX.exr}, representing the depth along the z-axis of the camera, from a minimum of 0.1 mm to a maximum of 200 mm. Depth values are linearly normalized between 0 and 1.
  \item Optical flow, \textit{Flow/Flow\_XXXX.exr}, representing the pixels' movement between the current, following, and previous frames as a four-channel image. The first two channels contain the X and Y components of optical flow, computed from the following frame to the current frame, whereas the last two represent the pixel displacement from the current to the previous frame.
  \item Trajectory,\textit{Trajectory.txt}, given as a text file where each line is associated with the frame's camera-to-world pose. Each line is composed by 13 elements: frame id (which refers to the frame number in the sequence), 3 elements for translation and 9 elements for the rotation matrix in row-major order.
  \item Camera intrinsic parameters, \textit{Intrinsic.txt}, given as a 3x3 standard intrinsic matrix for the pinhole model.
  \item 3D Mesh, \textit{Colon\_XX.ply}, containing the model ground truth mesh given as a .ply file.  
\end{itemize}

Depth maps and optical flows are saved as 16-bit OpenEXR images, compressed with PIZ lossless algorithm \cite{kainz2013}.

\begin{table}[t]
\centering
\caption{Details and structure of the proposed RealSynCol}
\label{tab:structure}
\resizebox{0.7\columnwidth}{!}{%
\begin{tabular}{ccccc}
\textbf{Geometry} & \textbf{Acquisition Mode} & \textbf{Trajectory} & \textbf{Step} & \textbf{Number of frames} \\ \hline
\multirow{2}{*}{1} & \multirow{2}{*}{Supine} & Backward & 5 & 1\,281 \\ \cline{3-5} 
& & Forward & 5 & 1\,281 \\ \hline
\multirow{2}{*}{2} & \multirow{2}{*}{Supine} & Backward & 2 & 3\,001 \\ \cline{3-5} 
& & Forward & 2 & 3\,001 \\ \hline
\multirow{2}{*}{3} & \multirow{2}{*}{Prone} & Backward & 5 & 1\,301 \\ \cline{3-5} 
& & Forward & 5 & 1\,301 \\ \hline
\multirow{2}{*}{4} & \multirow{2}{*}{Supine} & Backward & 2 & 3\,501 \\ \cline{3-5} 
& & Forward & 2 & 3\,501 \\ \hline
\multirow{2}{*}{5} & \multirow{2}{*}{Prone} & Backward & 5 & 1\,201 \\ \cline{3-5} 
& & Forward & 5 & 1\,201 \\ \hline
\multirow{2}{*}{6} & \multirow{2}{*}{Supine} & Backward & 10 & 561 \\ \cline{3-5} 
& & Forward & 10 & 561 \\ \hline
\multirow{2}{*}{7} & \multirow{2}{*}{Prone} & Backward & 10 & 556 \\ \cline{3-5} 
& & Forward & 10 & 556 \\ \hline
\multirow{2}{*}{8} & \multirow{2}{*}{Prone} & Backward & 5 & 1\,201 \\ \cline{3-5} 
& & Forward & 5 & 1\,201 \\ \hline
\multirow{2}{*}{9} & \multirow{2}{*}{Prone} & Backward & 5 & 861 \\ \cline{3-5} 
& & Forward & 5 & 861 \\ \hline
\multirow{2}{*}{10} & \multirow{2}{*}{Supine} & Backward & 10 & 601 \\ \cline{3-5} 
& & Forward & 10 & 601 \\ \hline
\end{tabular}%
}
\end{table}

\section{Technical Validation}
Different analyses were performed to thoroughly validate RealSynCol. First, a state-of-the-art depth foundation model, Depth Anything Model (DAM) v2 \cite{yang2024} was evaluated in a zero-shot setting and after fine-tuning on RealSynCol, to assess the relevance of high-quality synthetic dataset paired with ground truth information. 

Then, an ablation study was conducted to evaluate the impact of different dataset elements on the learning and generalization capabilities of deep learning algorithms, for both the tasks of depth and pose estimation. 

A benchmark study was also performed to compare public synthetic colon datasets for the task of self-supervised monocular depth and pose estimation. 

For the ablation study and the benchmark study, Lite-Mono~\cite{zhang2023}, a state-of-the-art self-supervised approach that jointly learns depth and ego-motion from pairs of consecutive frames, was employed. The depth estimation network includes a Consecutive Dilated Convolutions module to capture high-quality local features, and a Local-Global Features Interaction module (with multi-head self-attention blocks) to encode global contexts into the features.
Lite-Mono was chosen as the benchmark framework due to its strong methodological grounding, as it belongs to the family of self-supervised depth and motion estimation models originating from Monodepth2 \cite{monodepth2}. While models in this family may differ in the depth network, some terms of the loss functions, or auxiliary networks, they all share the same pose network and the general structure of the self-supervised training loss. Therefore, Lite-Mono provides a reliable reference for dataset benchmarking, while still being representative of the broader class of self-supervised depth and motion estimation models.

\subsection{Evaluation Metrics}\label{sec:eval_metrics}
The depth estimation performance is evaluated in terms of absolute relative error (Abs Rel), squared relative error (Sq Rel), Root Mean Square Error (RMSE), Root Mean Square Error in the logarithmic space (RMSE log), and Threshold Accuracy ($\delta$), defined in Table \ref{tab:depth_metrics}.

\begin{table}[!ht]
\centering
\caption{Evaluation metrics used for depth estimation. \( d \) is the estimated depth value, \( d^* \) is the ground truth depth value, and \( N \) is the total number of pixels. For the accuracy metric \( \delta \), \(\tau \in \{1.25,\; 1.25^2,\; 1.25^3\}\) denotes the used thresholds.}
\label{tab:depth_metrics}
\renewcommand{\arraystretch}{2}

\begin{tabularx}{0.60\linewidth}{@{} >{\centering\arraybackslash}X @{}}
\toprule
\textbf{Metric} \\
\midrule
Abs Rel = $\displaystyle \frac{1}{|N|} \sum_{i \in N} \frac{|y_i - \hat{y}_i|}{y_i}$ \\
Sq Rel = $\displaystyle \frac{1}{|N|} \sum_{i \in N} \frac{(y_i - \hat{y}_i)^2}{y_i}$ \\
RMSE = $\displaystyle \sqrt{\frac{1}{|N|} \sum_{i \in N} (y_i - \hat{y}_i)^2}$ \\
RMSE$_{\log}$ = $\displaystyle \sqrt{\frac{1}{|N|} \sum_{i \in N} (\log y_i - \log \hat{y}_i)^2}$ \\
$\delta$ = $\displaystyle \frac{1}{|D|} \left| \left\{ d \in D \;\middle|\; 
\max \left( \frac{d^*}{d}, \frac{d}{d^*} \right) < \tau \right\} \right| \times 100\%$ \\
\bottomrule
\end{tabularx}

\end{table}

To evaluate the pose estimation performance, following a protocol inspired by \cite{simcol}, the predicted relative poses are first combined to obtain the trajectory of absolute predicted poses. When dealing with predictions of neural networks on monocular sequences, this trajectory must be rescaled to match the scale of the ground truth trajectory. The scale factor was computed using the following formula:

\begin{equation}
\text{\textit{scale factor}} = \frac{\sum_{i} \mathbf{t}_i^{\text{gt}} \cdot \mathbf{t}_i^{\text{pred}}}{\sum_{i} \left\| \mathbf{t}_i^{\text{pred}} \right\|^2}
\end{equation}

where $\mathbf{t}_i^{\text{gt}}$ and $\mathbf{t}_i^{\text{pred}}$ are the translation vectors of the ground truth and predicted poses, respectively. The scaled predicted trajectory and the groundtruth trajectory were then aligned to make sure the first point was common. The pose estimation module was finally evaluated in terms of Absolute Trajectory Error (ATE), Relative Translation Error (RTE), Rotation Error (ROT), Root Mean Square Error for Translation (RMSE\_transl), and Root Mean Square Error for Rotation (RMSE\_rot). 

\begin{table}[!ht]
\centering
\caption{Metrics used to evaluate pose estimation.}
\label{tab:pose_metrics}
\renewcommand{\arraystretch}{2}

\begin{tabularx}{0.60\linewidth}{@{} >{\centering\arraybackslash}X @{}}
\toprule
\textbf{Metric} \\
\midrule
ATE = $\displaystyle \left\| \mathbf{t}_i^{\text{gt}} - \mathbf{t}_i^{\text{pred}} \right\|_2$ \\
RTE = $\displaystyle \left\| \mathbf{t}_{i+1}^{\text{gt}} - \mathbf{t}_i^{\text{gt}} -
\left( \mathbf{t}_{i+1}^{\text{pred}} - \mathbf{t}_i^{\text{pred}} \right) \right\|_2$ \\
ROT = $\displaystyle \arccos \left(
\frac{ \text{Tr}\left( \left(\mathbf{R}_i^{\text{gt}}\right)^{-1}
\mathbf{R}_i^{\text{pred}} \right) - 1 }{2} \right)$ \\
RMSE$_\text{transl}$ = $\displaystyle
\sqrt{ \frac{1}{N} \sum_{i=1}^{N}
\left\| \mathbf{t}_i^{\text{gt}} - \mathbf{t}_i^{\text{pred}} \right\|_2^2 }$ \\
RMSE$_\text{rot}$ = $\displaystyle
\sqrt{ \frac{1}{N-1} \sum_{i=1}^{N-1}
\left( \text{ROT}_i \right)^2 }$ \\
\bottomrule
\end{tabularx}

\end{table}

\subsection{Training Parameters}\label{sec:training_parameters}
All the trainings were conducted in Pytorch on a an NVIDIA A100 GPU. 
The DAM v2 small was finetuned starting from the weights from the original repository \cite{yang2024}, and using images at 518x518 pixel resolution. The optimizer is AdamW, the weight decay is $1\mathrm{e}{-2}$, and the initial learning rate is $1\mathrm{e}{-4}$, decreasing with a cosine rate schedule.

For all experiments conducted with Lite-Mono, images were downsampled at 512x512 pixels resolution with bicubic interpolation and fed to the network. The pose estimation network is based on a ResNet encoder with 18 layers. Following the hyperparameters in the official repository \cite{zhang2023}, the selected optimizer is AdamW, the weight decay is $1\mathrm{e}{-2}$, whereas the initial learning rate is $1\mathrm{e}{-4}$, decreasing with a cosine rate schedule. The network is trained for a total of 50 epochs, and the final weights are used for the pose estimation task. For the depth estimation task, the best performance is achieved after 20 epochs on average, so these checkpoints are used for testing.

\subsection{Impact on zero-shot vs fine-tuned metric depth estimation}

Recent progress in computer vision has been driven by the development of foundation models, which exhibit strong cross-domain generalization as a result of extensive pre-training on heterogeneous data. For depth estimation task, these models typically predict relative or scale-ambiguous depth, and need to be adapted to achieve accurate and meaningful metric depth estimation. This explicit fine-tuning requires dense and reliable metric ground truth. Such supervision, however, is unfeasible to obtain from real clinical endoscopic data, due to the lack of direct depth sensing and the impracticality of acquiring precise geometric annotations in vivo. This limitation motivates the use of high-quality synthetic datasets with accurate metric annotations. 

Since this work aims to establish RealSynCol as a benchmark for endoscopic 3D reconstruction, we explicitly investigate whether and to what extent a depth foundation model benefits from fine-tuning on such synthetic data, by comparing its zero-shot performance against its performance after adaptation to the proposed dataset. We evaluate the impact of RealSynCol on the adaptation of a recent state-of-the-art depth foundation model, DAM v2 \cite{yang2024}, originally trained on large-scale datasets spanning multiple visual domains but not including endoscopic imaging. The goal of this experiment is to explicitly assess whether access to high-quality synthetic supervision enables a foundation model to transition from relative to accurate metric depth estimation in endoscopy. To this end, we consider two configurations: zero-shot DAM v2, where the model is evaluated using the publicly released weights without any endoscopy-specific adaptation, and fine-tuned DAM v2, where the same model is adapted using the RealSynCol training set.

For this experiment, \textit{Colon\_9\_backward} and \textit{Colon\_9\_forward} are reserved as the test set, while all remaining sequences are used for training and validation. We fine-tune the DAM v2 small variant for 10 epochs using Low-Rank Adaptation (LoRA) \cite{hu2022} with rank 8 and an input resolution of $518 \times 518$. Following standard practice for adapting vision transformers, LoRA modules are injected into all linear mappings within the transformer encoder blocks, including both self-attention projections and feed-forward sublayers.

Evaluation is performed using the depth estimation metrics described in Sec.~\ref{sec:eval_metrics}. In the zero-shot configuration, DAM v2 predicts depth maps only up to an unknown global scale. To enable quantitative comparison, we apply median scaling with respect to the ground-truth depth maps, following the standard protocol adopted in monocular depth estimation benchmarks \cite{monodepth2}. In contrast, after fine-tuning on RealSynCol, the model directly outputs depth maps in metric scale, without requiring any test-time scale alignment.

\begin{figure} 
\centering 
\includegraphics[width=0.6\columnwidth]{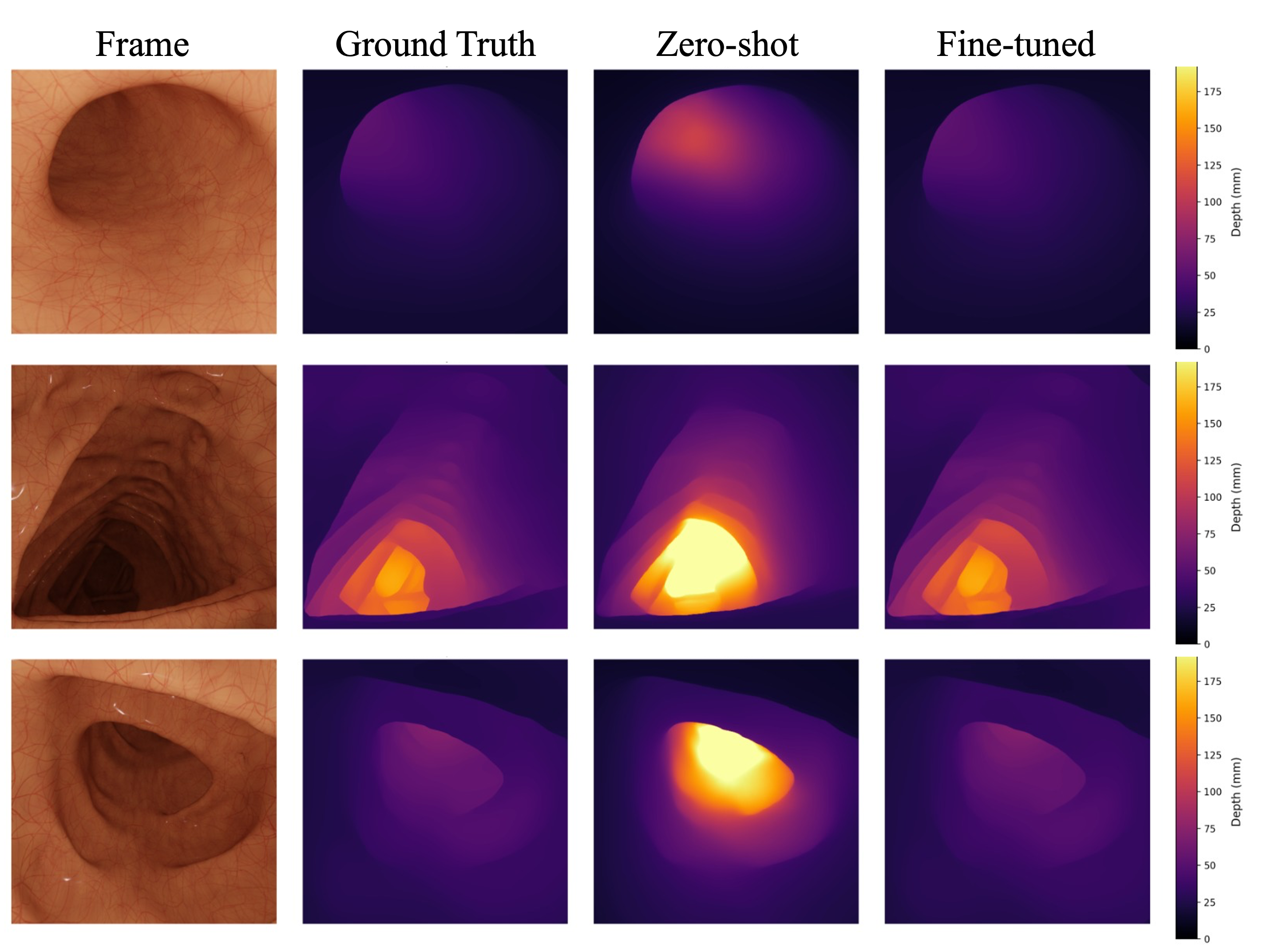}
\caption{Examples of depth maps predicted using zero-shot DAM v2 and fine-tuned DAM v2 }
\label{fig:foundation_models}
\end{figure}

Although DAM v2 is formulated to regress disparity rather than depth, the availability of dense and accurate metric annotations in RealSynCol enables the network to learn a consistent mapping from disparity to absolute depth during fine-tuning. Quantitative results comparing the zero-shot and fine-tuned configurations are reported in Tab.~\ref{tab:foundation_models}, and qualitative examples are shown in Fig.~\ref{fig:foundation_models}.

The results clearly indicate that, while the zero-shot foundation model produces visually plausible depth maps, its predictions lack metric consistency and require post-hoc scale correction. Fine-tuning on RealSynCol leads to a substantial improvement in metric accuracy and geometric fidelity, and crucially removes the need for any test-time scale adjustment, which is infeasible in real clinical scenarios. This experiment directly demonstrates that, even in the presence of powerful and robust foundation models, effective endoscopic 3D reconstruction critically depends on access to high-quality synthetic datasets providing dense and reliable metric ground truth. In this sense, RealSynCol serves not only as a benchmarking resource, but also as a necessary enabler for adapting modern foundation models to the specific geometric and visual characteristics of endoscopic imaging.

\begin{table}[!ht]
\centering
\caption{Quantitative comparison between zero-shot and fine-tuned DAM v2 for depth estimation on RealSynCol test set in terms of Abs Rel, Sq Rel, RMSE, RMSE log, and $\delta$ metrics with \(\tau \in \{1.25, 1.25^2, 1.25^3\}\). Best results are highlighted in bold. Predictions obtained with zero-shot DAM v2 are rescaled using median scaling to enable quantitative evaluation.}
\label{tab:foundation_models}
\resizebox{\textwidth}{!}{%
\begin{tabular}{l
                r@{ $\pm$ }l
                r@{ $\pm$ }l
                r@{ $\pm$ }l
                r@{ $\pm$ }l
                r@{ $\pm$ }l
                r@{ $\pm$ }l
                r@{ $\pm$ }l}
\hline
\textbf{Model} &
\multicolumn{2}{c}{\textbf{Abs rel} $\downarrow$} &
\multicolumn{2}{c}{\textbf{Sq rel} $\downarrow$} &
\multicolumn{2}{c}{\textbf{RMSE} $\downarrow$} &
\multicolumn{2}{c}{\textbf{RMSE log} $\downarrow$} &
\multicolumn{2}{c}{\textbf{$\delta < 1.25$} $\uparrow$} &
\multicolumn{2}{c}{\textbf{$\delta < 1.25^2$} $\uparrow$} &
\multicolumn{2}{c}{\textbf{$\delta < 1.25^3$} $\uparrow$} \\
\hline
Zero-shot DAM v2 &
0.1641 & 0.0347 &
1.8232 & 0.8106 &
8.9968 & 3.2165 &
0.2915 & 0.0682 &
0.7126 & 0.0904 &
0.8901 & 0.0562 &
0.9539 & 0.0314 \\
Fine-tuned DAM v2 &
\textbf{0.0168} & \textbf{0.0057} &
\textbf{0.0210} & \textbf{0.0142} &
\textbf{0.9041} & \textbf{0.3862} &
\textbf{0.0228} & \textbf{0.0074} &
\textbf{0.9998} & \textbf{0.0002} &
\textbf{1.0000} & \textbf{0.0001} &
\textbf{1.0000} & \textbf{0.0000} \\
\hline
\end{tabular}%
}
\end{table}

\subsection{Ablation Study}\label{sec:ablation_study}
For this study, two alternative versions of the RealSynCol dataset were generated to evaluate the impact of the realistic texture and light reflections on the performance of the self-supervised approach. In the first version, the realistic colon texture was replaced with a uniform base color, sampled from clinical images from the SUN Database~\cite{SUN}. In the second version, the dataset was created using the realistic texture generated with the protocol described in Sec.~\ref{sec:virtual_env}, but no specular reflections were added during the rendering phase. The performance obtained with these two variants is then compared against the final version of the RealSynCol dataset. During the training process, for all dataset versions, \textit{Colon\_9\_backward} and \textit{Colon\_9\_forward} are designated as the test set, whereas the remaining models are used for training and validation. The three dataset versions are referred to as "Base Color", "Base Color + Texture", "Base Color + Texture + Reflections".

\begin{figure} 
\centering 
\includegraphics[width=0.4\columnwidth]{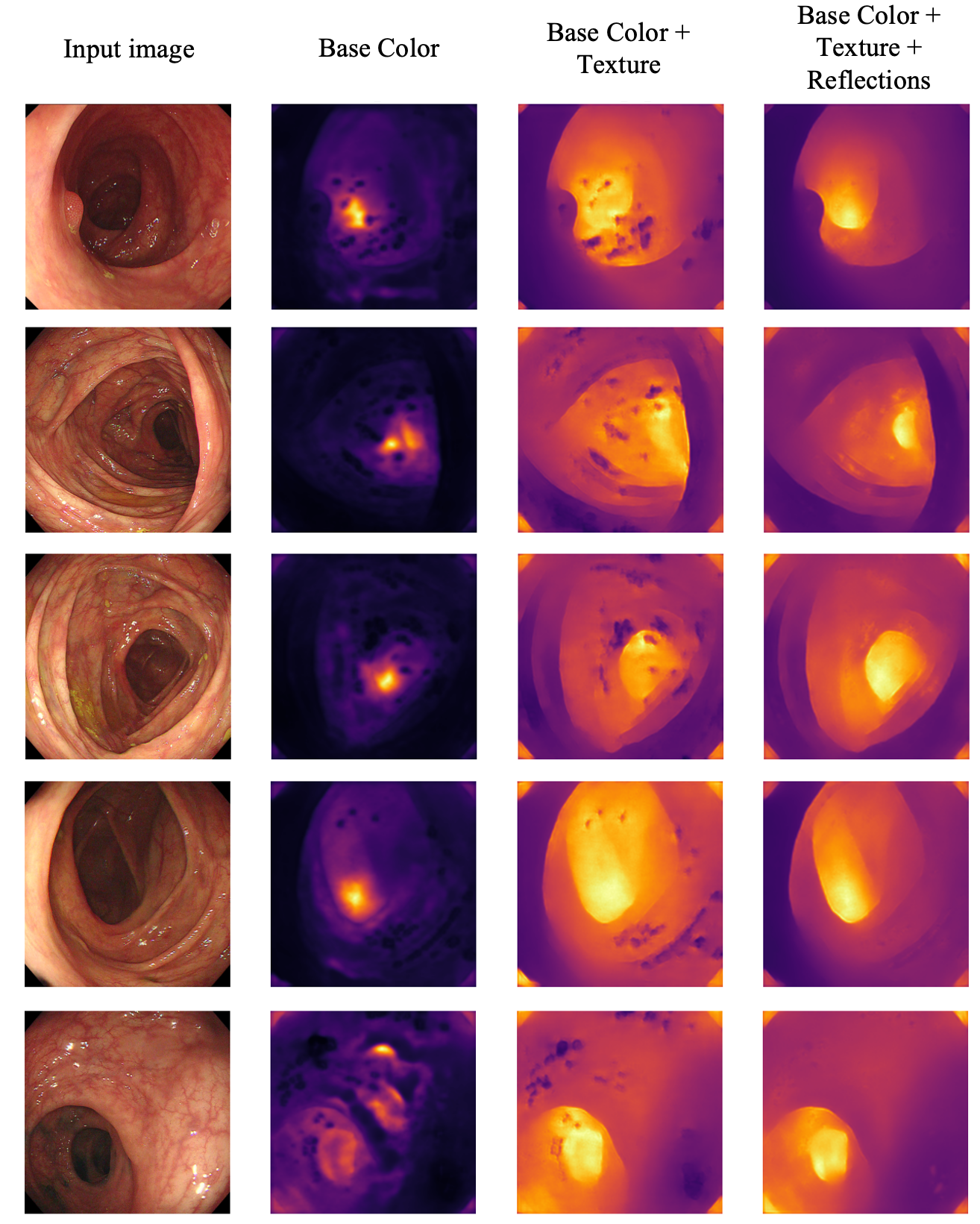}
\caption{Examples of depth maps from the clinical SUN Database \cite{SUN} predicted using the models included in the ablation study. For clinical images ground truth depth maps are not available. }
\label{fig:clinical_ablation_study_depth}
\end{figure}

Table~\ref{tab:ablation_study_depth} reports the quantitative metrics for depth estimation, whereas Fig.\ref{fig:clinical_ablation_study_depth} illustrates examples of predicted depth maps obtained using the models analyzed in the ablation study, based on clinical images from the SUN Database\cite{SUN}.

\begin{table*}[t]
\caption{Ablation study results for depth estimation task in terms of Abs Rel, Sq Rel, RMSE, RMSE log, and $\delta$ metrics with \(\tau \in \{1.25, 1.25^2, 1.25^3\}\). Best results are highlighted in bold.}
\label{tab:ablation_study_depth}
\resizebox{\textwidth}{!}{%
\begin{tabular}{l
                r@{ $\pm$ }l
                r@{ $\pm$ }l
                r@{ $\pm$ }l
                r@{ $\pm$ }l
                r@{ $\pm$ }l
                r@{ $\pm$ }l
                r@{ $\pm$ }l}
\hline
\textbf{Experiment} 
& \multicolumn{2}{c}{\textbf{Abs Rel} $\downarrow$} 
& \multicolumn{2}{c}{\textbf{Sq Rel} $\downarrow$} 
& \multicolumn{2}{c}{\textbf{RMSE} $\downarrow$} 
& \multicolumn{2}{c}{\textbf{RMSE log} $\downarrow$} 
& \multicolumn{2}{c}{\textbf{$\delta < 1.25$} $\uparrow$} 
& \multicolumn{2}{c}{\textbf{$\delta < 1.25^2$} $\uparrow$} 
& \multicolumn{2}{c}{\textbf{$\delta < 1.25^3$} $\uparrow$} \\ 
\hline
Base color 
& 2.564 & 0.799 
& 379.232 & 149.976 
& 101.321 & 18.758 
& 1.621 & 0.605 
& 0.069 & 0.065 
& 0.147 & 0.113 
& 0.240 & 0.158 \\

Base Color + Texture 
& \textbf{0.098} & \textbf{0.017} 
& \textbf{0.725} & \textbf{0.412} 
& \textbf{5.705} & \textbf{2.049} 
& \textbf{0.134} & \textbf{0.031} 
& \textbf{0.931} & \textbf{0.049} 
& \textbf{0.992} & \textbf{0.010} 
& \textbf{0.997} & \textbf{0.007} \\

Base Color + Texture + Reflections 
& 0.100 & 0.016 
& 0.773 & 0.404 
& 5.996 & 2.079 
& 0.136 & 0.030 
& 0.926 & 0.045 
& \textbf{0.992} & \textbf{0.010} 
& \textbf{0.997} & \textbf{0.007} \\ 
\hline
\end{tabular}%
}
\end{table*}

The model trained on the dataset without texture (i.e., using only a uniform base color) exhibits substantially higher error values, highlighting how the absence of realistic mucosal patterns impairs the accurate learning of depth and motion cues. Training with and without light reflections yields comparable results, with slightly lower errors observed in the model trained on images without reflections. This indicates that incorporating a realistic light reflection model slightly increases task difficulty during the learning phase. However, the analysis of clinical depth maps in Fig.~\ref{fig:clinical_ablation_study_depth} clearly demonstrates that a more realistic training dataset produces more accurate and stable predictions, with fewer artifacts in correspondence with specular reflections or highly saturated regions.

\begin{figure} 
\centering 
\includegraphics[width=0.5\columnwidth]{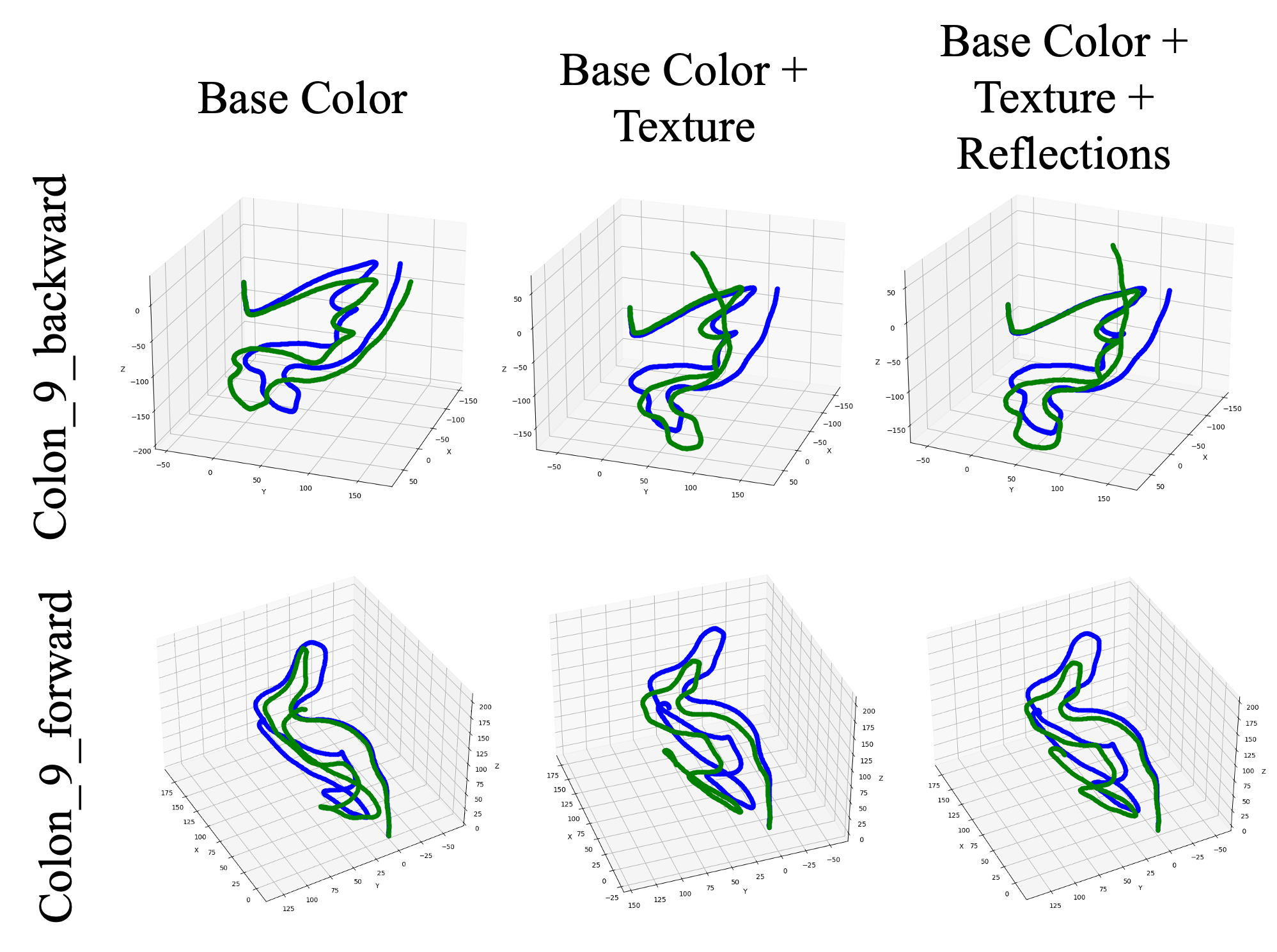}
\caption{Predicted trajectories for Colon 9 (forward and backward) using the models from the ablation study. Ground truth trajectories are shown in blue, and predicted trajectories in green.}
\label{fig:trajectories_ablation_study}
\end{figure}

In Table~\ref{tab:ablation_study_pose}, pose estimation metrics are reported for the two colon trajectories in the test set (\textit{Colon\_9\_forward} and \textit{backward}) and the two trajectories in the validation set (\textit{Colon\_5\_forward} and \textit{backward}). The inclusion of more realistic surface textures and reflections results in improved accuracy in the estimation of rotational motions. The analysis of translational movements shows consistently strong performance across all dataset variants, although some variability in movement patterns is observed. Specifically, translation accuracy improves in the two validation sequences and in the backward trajectory of the test set, while a slight degradation is observed in the forward trajectory of the test set. In the \textit{Colon\_9\_forward} trajectory, the first colonic bend is particularly sharp and narrow. As a result, several consecutive frames lose sight of the lumen entirely, reducing spatial awareness as the camera captures only nearby tissue without any background reference. Under these conditions, where spatial cues are limited, the presence of vascular patterns may slightly increase the complexity of motion estimation. Consequently, texture-based models tend to accumulate a minor drift in this segment, increasing the translational error. Nonetheless, beyond this region, the overall reconstruction of the three-dimensional trajectory remains highly accurate, as pictured in Fig.~\ref{fig:trajectories_ablation_study}.

\begin{table*}[t]
\caption{Ablation study results for pose estimation task in terms of ATE, RTE, ROT, RMSE transl, RMSE rot. ATE, RTE, and ROT are reported as median values, and the interquartile range is in brackets. To provide a comprehensive overview of performance, metrics are reported for both validation sequences (Colon 5 forward and backward) and test sequences (Colon 9 forward and backward). Best results for every sequence are highlighted in bold.}
\label{tab:ablation_study_pose}

\resizebox{\textwidth}{!}{
\begin{tabular}{l c c c c c c}
\hline
\textbf{Experiment} & \textbf{Sequence}  & \textbf{ATE (mm)} $\downarrow$ & \textbf{RTE (mm)} $\downarrow$ & \textbf{ROT (deg)} $\downarrow$ & \textbf{RMSE transl (mm)} $\downarrow$ & \textbf{RMSE rot (deg)} $\downarrow$ \\ \hline
Base Color          & Colon\_5\_backward & 69.081(59.292)   & 0.075 (0.053)     & 0.121 (0.095)      & 82.140                    & 0.187                  \\
Base Color + Texture      & Colon\_5\_backward & 34.628 (41.421)   & 0.065 (0.056)     & 0.079 (0.060)      & 50.736                    & 0.099                  \\
Base Color + Texture + Reflections       & Colon\_5\_backward & \textbf{29.992 (20.119)}   & \textbf{0.064 (0.045)}     & \textbf{0.072 (0.054)}      & \textbf{40.905}                    & \textbf{0.093}                  \\ \hline
Base Color          & Colon\_5\_forward & 61.103(84.292)   & 0.094 (0.060)     & 0.120 (0.100)      & 85.321                   & 0.188                  \\
Base Color + Texture      & Colon\_5\_forward & 28.386 (27.602)   & \textbf{0.064 (0.045)}     & 0.076 (0.057)      & 39.243                    & 0.097                  \\
Base Color + Texture + Reflections       & Colon\_5\_forward & \textbf{24.268 (24.526)}   & 0.065 (0.042)     & \textbf{0.072 (0.052)}      & \textbf{36.912}                    & \textbf{0.091}                  \\ \hline
Base Color          & Colon\_9\_backward & 28.659 (15.854)   & \textbf{0.165 (0.134)}     & 0.278 (0.308)      & \textbf{31.027}                    & 0.748                  \\
Base Color + Texture      & Colon\_9\_backward & 23.202 (26.594)   & 0.175 (0.111)     & 0.192 (0.198)      & 46.273                    & 0.694                  \\
Base Color + Texture + Reflections       & Colon\_9\_backward & \textbf{20.532 (18.853)}   & 0.179 (0.151)     & \textbf{0.169 (0.146)}      & 37.677                    & \textbf{0.600}                  \\ \hline
Base Color          & Colon\_9\_forward  & \textbf{14.746 (12.548)}   & \textbf{0.152 (0.120)}     & 0.264 (0.331)      & \textbf{17.784}                    & 0.727                  \\
Base Color + Texture      & Colon\_9\_forward  & 33.791 (40.908)   & 0.259 (0.170)     & 0.194 (0.193)      & 47.772                    & 0.535                  \\
Base Color + Texture + Reflections       & Colon\_9\_forward  & 33.567 (37.402)   & 0.231 (0.153)     & \textbf{0.168 (0.149)}      & 40.948                    & \textbf{0.461}                 
\end{tabular}%
}
\end{table*}

\subsection{Benchmark analysis}\label{sec:benchmark_analysis}
A benchmark study was performed to compare public synthetic colon datasets for the task of monocular depth estimation and pose estimation, using Lite-Mono~\cite{zhang2023}, a self-supervised approach that jointly learns depth and ego-motion from pairs of consecutive frames. 

\subsubsection{Data Preparation}\label{sec:data_preparation}
\begin{figure} 
\centering 
\includegraphics[width=0.5\columnwidth]{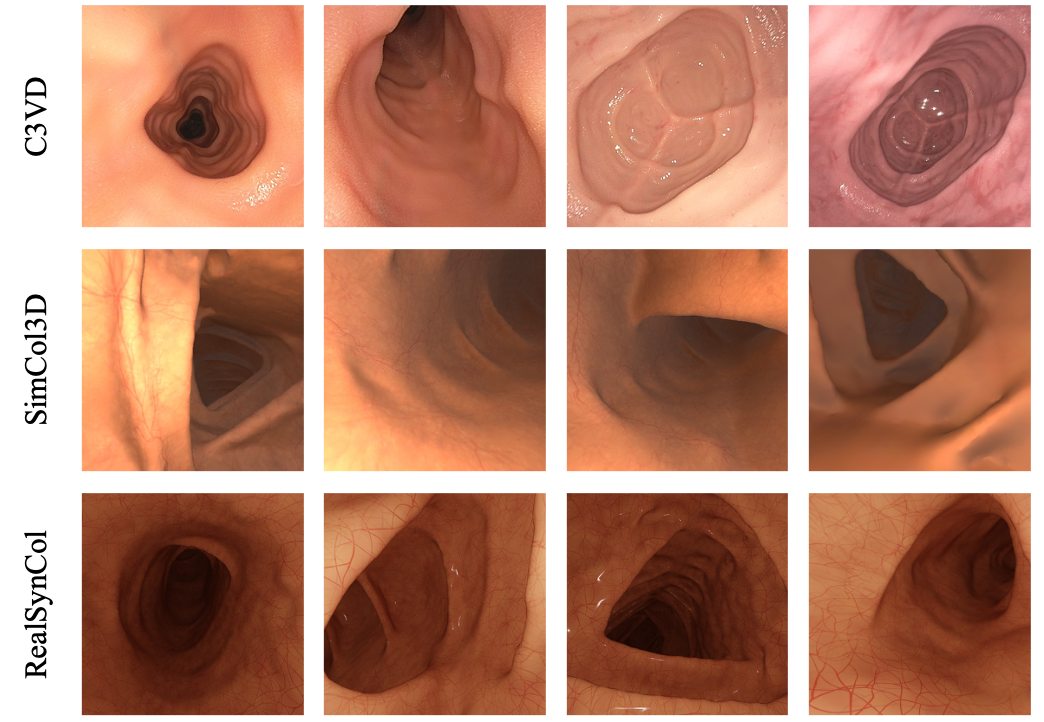}
\caption{Example of images from C3VD, SimCol3D, and RealSynCol.}
\label{fig:examples_datasets}
\end{figure}

RealSynCol was evaluated against SimCol3D~\cite{simcol} and C3VD~\cite{C3VD}. EndoSLAM~\cite{EndoSLAM} was excluded as all images come from the same sequence, so there is no variability in colon trajectories. Endomapper~\cite{Endomapper} was excluded due to the very limited number of images and the low variability among the sequences. 
We also did not include the dataset in~\cite{Rau2019}, as the colon model used to generate the frames was included in the bigger and more recent SimCol3D. Examples of images from C3VD, SimCol3D, and RealSynCol are depicted in Fig.\ref {fig:examples_datasets}.

To better clarify the characteristics of these datasets, an analysis of the relative movements and rotations between consecutive frames was conducted. To ensure consistency, given that the three datasets were generated using different software with potentially non-aligned axis orientations, relative displacements were computed as the norm of translations along the three axes (x, y, and z). The results of this analysis are presented in Table \ref{tab:relative_translations}. The first row of the table, referred to as "Phantom Acquisition", reports the median relative displacement obtained from the laboratory acquisition by an expert clinician on a deformable phantom with the protocol described in Sec. \ref{sec:virtual_env}.
The analysis clearly shows a marked difference in the range of motion among the three datasets: in C3VD, consecutive frames are almost static, whereas the other two datasets display considerably larger movements. In RealSynCol, the variability in motion magnitude reflects a broader range of relative motions in the proposed dataset, better capturing the clinical scenario in which the endoscope is manually operated and thus exhibits more complex displacement patterns. Both RealSynCol and SimCol3D exhibit a median range of movement comparable to that measured in phantom acquisitions, while the variability in displacement magnitude is higher for RealSynCol, consistent with the elevated interquartile range observed in the phantom data.

\begin{figure} 
\centering 
\includegraphics[width=0.5\columnwidth]{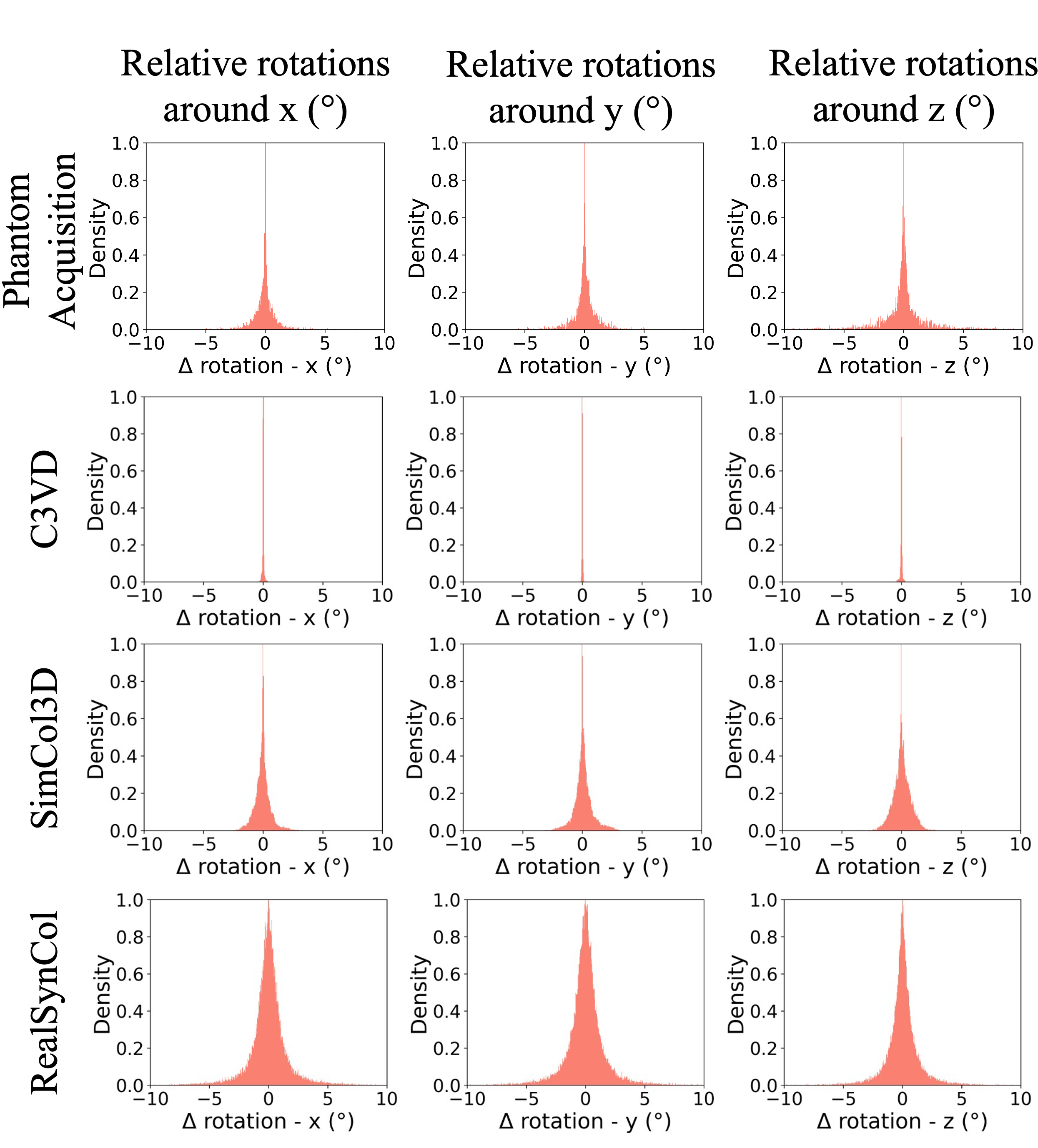}
\caption{Analysis of relative rotations around x, y, and z axes between consecutive frames in Phantom Acquisition (endoscope moved by an experienced clinician in a silicone colon phantom), C3VD, SimCol3D, RealSynCol. All rotations are represented in degrees.}
\label{fig:hist_rotations}
\end{figure}

An analysis of the relative rotations is pictured in Fig.\ref{fig:hist_rotations}. C3VD shows the most static frames in terms of relative rotation. Although all datasets show rotation values predominantly centered around zero, RealSynCol exhibits a significantly broader range of rotational motions. 
As described in Sec.~\ref{sec:virtual_env}, this analysis also indicates that rotations in RealSynCol are slightly larger than those observed during acquisitions on the phantom setup performed by an experienced clinician. Colon phantoms exhibit less pronounced haustral folds and gentler curvatures compared to real anatomy. Consequently, when navigating the phantom with the objective of visualizing the largest possible mucosal area, as in standard colonoscopy, the endoscope rotations tend to be less pronounced than those typically observed in clinical practice, as also noted by the endoscopists involved in the acquisition. Therefore, in RealSynCol, the parameters controlling orientation noise were tuned to produce slightly larger rotations, to account for challenging navigation scenarios, such as sharp bends or lesions located along the borders of haustral folds.

It is also worth noting that some sequences in C3VD exhibit both forward and backward movements along the camera axis. In RealSynCol, each colon model includes both a forward and a backward trajectory, whereas all sequences in SimCol3D feature forward motion only. Regarding variability in colon shapes and textures, SimCol3D comprises images from three distinct colon shapes, all rendered with the same texture; C3VD employs a single colon shape with four different textures; and RealSynCol was generated using ten different colon shapes with a single texture.

\begin{table}[t]
\centering
\caption{Analysis of relative movements between consecutive frames in Phantom Acquisition (endoscope moved by an experienced clinician in a silicone colon phantom), C3VD, SimCol3D and RealSynCol. Relative movements are computed as the norm of translations along the three axes, x, y, and z. The reported results are the median values in mm with interquartile range in brackets.}
\label{tab:relative_translations}
\resizebox{0.65\columnwidth}{!}{%
\begin{tabular}{lcc}
\textbf{Dataset} & \multicolumn{1}{c}{\textbf{Number of frames}} & \multicolumn{1}{c}{\textbf{Displacement}} \\ \hline
Phantom Acquisition &  3\,213 & 0.81 (1.30) \\ \hline
C3VD & 10\,015 & 0.06 (0.09) \\ \hline
SimCol3D & 23\,421 & 0.86 (0.37) \\ \hline
RealSynCol & 28\,130 & 0.90 (0.63) \\ \hline

\end{tabular}%
}
\end{table}


The input of the network is composed of pairs of consecutive frames and the camera intrinsic matrix. In RealSynCol and SimCol3D, the camera model is a pinhole, and the intrinsic matrix is available, whereas C3VD uses an omnidirectional sixth-grade polynomial camera model, where the standard intrinsic matrix is not available. C3VD images were undistorted using the procedure described in \cite{scaramuzza2006flexible, rufli2008}, and the parameters identified using the MATLAB toolbox OCamCalib \cite{scaramuzza2006toolbox}. The undistorted calibration images were then used to calibrate a pinhole camera model to obtain the intrinsic matrix. Undistorted images from C3VD were finally cropped to make them square, and for a fair comparison, all images were reshaped to a 512x512 pixel resolution.

Each of the considered datasets was split into training, validation, and test (approximately 80/10/10 of the total frames).
For C3VD, the sequences \textit{cecum\_t2\_a}, \textit{desc\_t4\_a}, \textit{sigmoid\_t3\_a}, and \textit{trans\_t3\_b} are used for testing, whereas the remaining sequences serve for training and validation. In SimCol3D, the test set includes sequences \textit{O1}, \textit{O2}, and \textit{O3}. For RealSynCol, \textit{Colon\_9\_backward} and \textit{Colon\_9\_forward} are used as the test set, while the other sequences are used for training and validation.
Since C3VD sequences are generally shorter than those in the other datasets, and only a single short sequence originates from a colon model excluded from training, we additionally evaluated all models on a full-length colon sequence, \textit{seq1}, released with the C3VD dataset. This allowed the assessment of the generalization capabilities over entire trajectories from previously unseen colons, rather than on short segments.

\begin{table*}[!ht]
\centering
\caption{Depth Estimation Results in terms of Abs Rel, Sq Rel, RMSE, RMSE log,  $\delta$ with \(\tau \in \{1.25,\; 1.25^2,\; 1.25^3\}\). For each dataset, the best metrics are in bold.}
\label{tab:depth_results}
\resizebox{\textwidth}{!}{%
\begin{tabular}{llccr@{\hspace{0.1cm}}c@{\hspace{0.1cm}}lccccl}
\toprule
\textbf{Dataset} & \textbf{Model trained on} & 
\textbf{Abs Rel } $\downarrow$ & \textbf{Sq Rel} $\downarrow$ & \multicolumn{3}{c}{\textbf{RMSE (mm)} $\downarrow$} & \textbf{RMSE log} $\downarrow$ &
$\boldsymbol{\delta < 1.25}$ $\uparrow$ & $\boldsymbol{\delta < 1.25^2}$ $\uparrow$ & $\boldsymbol{\delta < 1.25^3}$ $\uparrow$ \\
\midrule
C3VD & C3VD & 0.323 $\pm$ 0.171 & 4.531 $\pm$ 3.290 & 11.679 & $\pm$ & 3.972 & 0.379 $\pm$ 0.137 & 0.496 $\pm$ 0.201 & 0.780 $\pm$ 0.184 & 0.905 $\pm$ 0.112 \\
C3VD & SimCol3D & 0.155 $\pm$ 0.035 & 0.941 $\pm$ 0.400 & 5.726 & $\pm$ & 1.801 & 0.195 $\pm$ 0.043 & 0.749 $\pm$ 0.101 & 0.976 $\pm$ 0.035 & 0.993 $\pm$ 0.015 \\
C3VD & RealSynCol & \textbf{0.134} $\pm$ \textbf{0.045} & \textbf{0.700} $\pm$ \textbf{0.388} & \textbf{4.644} & $\pm$ & \textbf{1.684} & \textbf{0.164} $\pm$ \textbf{0.050} & \textbf{0.827} $\pm$ \textbf{0.113} & \textbf{0.986} $\pm$ \textbf{0.027} & \textbf{0.998} $\pm$ \textbf{0.011} \\
\midrule
SimCol3D & C3VD & 0.369 $\pm$ 0.097 & 6.076 $\pm$ 2.661 & 15.773 & $\pm$ & 4.064 & 0.450 $\pm$ 0.086 & 0.349 $\pm$ 0.102 & 0.669 $\pm$ 0.110 & 0.857 $\pm$ 0.087 \\
SimCol3D & SimCol3D & \textbf{0.058} $\pm$ \textbf{0.027} & \textbf{0.379} $\pm$ \textbf{1.030} & \textbf{3.477} & $\pm$ & \textbf{1.427} & \textbf{0.089} $\pm$ \textbf{0.033} & \textbf{0.977} $\pm$ \textbf{0.027} & \textbf{0.994} $\pm$ \textbf{0.007} & \textbf{0.997} $\pm$ \textbf{0.004} \\
SimCol3D & RealSynCol & 0.131 $\pm$ 0.050 & 1.492 $\pm$ 2.086 & 6.964 & $\pm$ & 3.217 & 0.166 $\pm$ 0.050 & 0.853 $\pm$ 0.115 & 0.980 $\pm$ 0.034 & 0.995 $\pm$ 0.007 \\
\midrule
RealSynCol & C3VD & 0.205 $\pm$ 0.053 & 3.500 $\pm$ 1.916 & 13.901 & $\pm$ & 5.714 & 0.311 $\pm$ 0.086 & 0.649 $\pm$ 0.129 & 0.868 $\pm$ 0.066 & 0.941 $\pm$ 0.051 \\
RealSynCol & SimCol3D & 0.119 $\pm$ 0.026 & 0.981 $\pm$ 0.466 & 6.750 & $\pm$ & 2.059 & 0.162 $\pm$ 0.040 & 0.866 $\pm$ 0.078 & 0.984 $\pm$ 0.017 & 0.996 $\pm$ 0.007 \\
RealSynCol & RealSynCol & \textbf{0.100} $\pm$ \textbf{0.016} & \textbf{0.773} $\pm$ \textbf{0.404} & \textbf{5.996} & $\pm$ & \textbf{2.079} & \textbf{0.136} $\pm$ \textbf{0.030} & \textbf{0.926} $\pm$ \textbf{0.045} & \textbf{0.992} $\pm$ \textbf{0.010} & \textbf{0.997} $\pm$ \textbf{0.007} \\
\bottomrule
\end{tabular}
}
\end{table*}

To assess the generalization capabilities associated with each dataset, we conducted a cross-dataset validation. Models trained on one dataset were evaluated on the remaining two to measure their performance on images with varying visual characteristics, illumination conditions, and textures. 

\begin{figure} [!ht]
\centering 
\includegraphics[width=0.5\columnwidth]{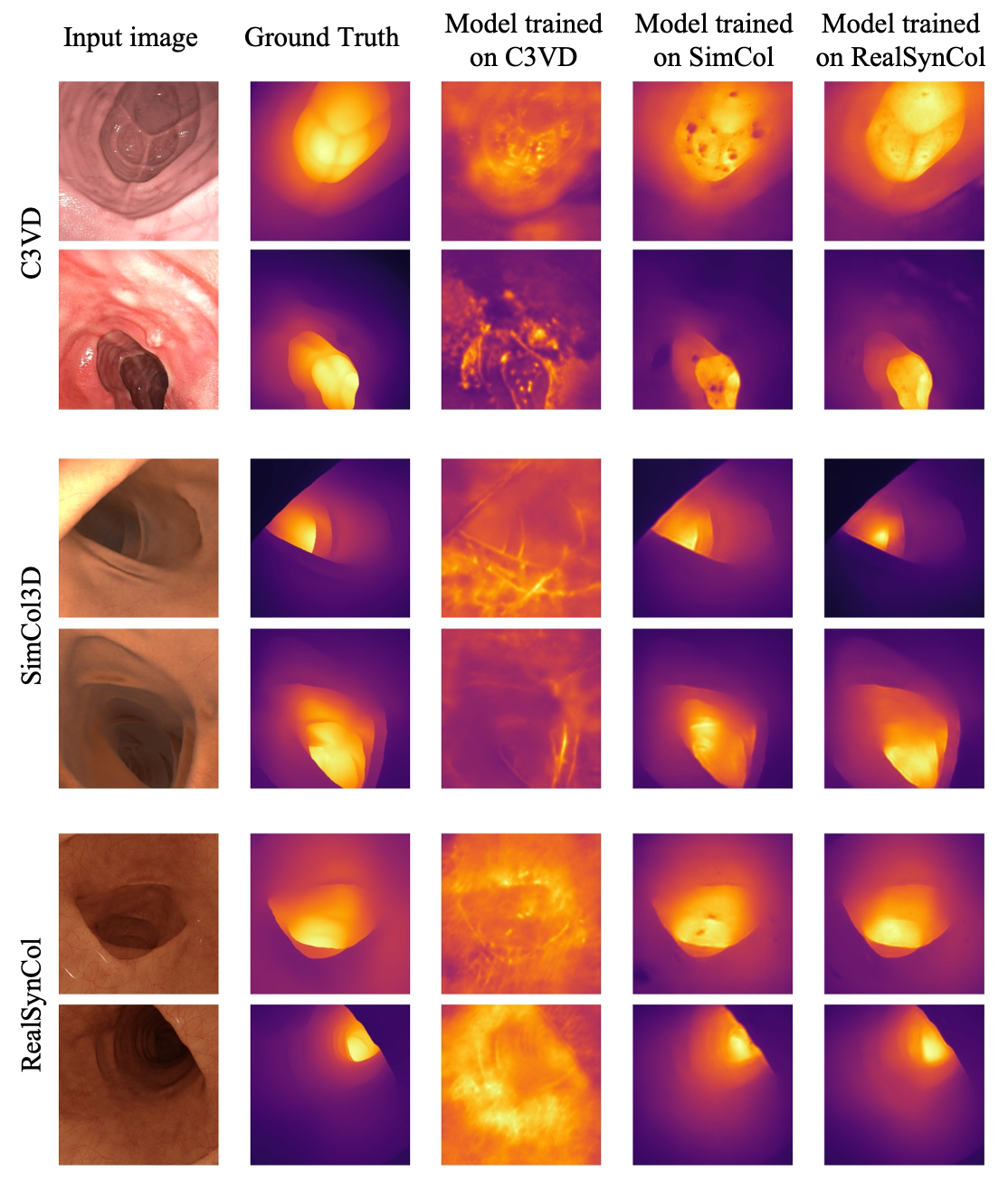}
\caption{Examples of depth maps predicted using the benchmark method Lite-Mono. The first two columns contain the input images and the relative ground truth depth maps, the following three columns contain the predictions computed with the model trained on C3VD, SimCol3D, and RealSynCol, respectively.} 
\label{fig:results_depth}
\end{figure}

\subsubsection{Depth estimation}
Table~\ref{tab:depth_results} reports depth estimation results using the evaluation metrics described in Sec.~\ref{sec:eval_metrics}. Examples of the predicted depth maps are shown in Fig~\ref{fig:results_depth}. 
Depth predictions obtained with models trained on SimCol3D and RealSynCol are generally robust and accurate, even when evaluated on datasets different from those used for training, demonstrating strong generalization capabilities. 
On C3VD images, the worst performance is observed with the model trained on C3VD itself, whereas the best results are obtained using the model trained on the proposed RealSynCol, and very good performance is achieved by the model trained on SimCol3D. As expected, on larger datasets that offer broader motion representation, the best results from the self-supervised model occur when training and testing are conducted on the same dataset. Cross-dataset evaluations between SimCol3D and RealSynCol yield comparable error levels, which remain acceptable within the depth range of 0.1–200 mm. However, for SimCol3D images, the predictions made using the model trained on RealSynCol exhibit an RMSE value approximately double that of the error obtained with the model trained on SimCol3D. This discrepancy may arise from overfitting of the model to the SimCol3D images, as many sequences are derived from the same colon model. The qualitative analysis highlights that predictions from the C3VD-trained model tend to be noisy and unstable, whereas those from the SimCol3D and RealSynCol models appear comparable. Notably, the RealSynCol-trained model demonstrates increased robustness to light reflections, likely a result of their explicit inclusion in the training set. A statistical analysis was conducted to assess the significance of the depth estimation results. For each of the three test subsets (C3VD, SimCol3D, and RealSynCol), the RMSE metric values obtained using the models trained on the three datasets were analyzed using the non-parametric Wilcoxon signed-rank test with Bonferroni correction. This analysis revealed a statistically significant difference in the results produced by the models trained on the different datasets across all three test subsets.

\begin{figure}[!htbp]
\centering 
\includegraphics[width=0.5\columnwidth]{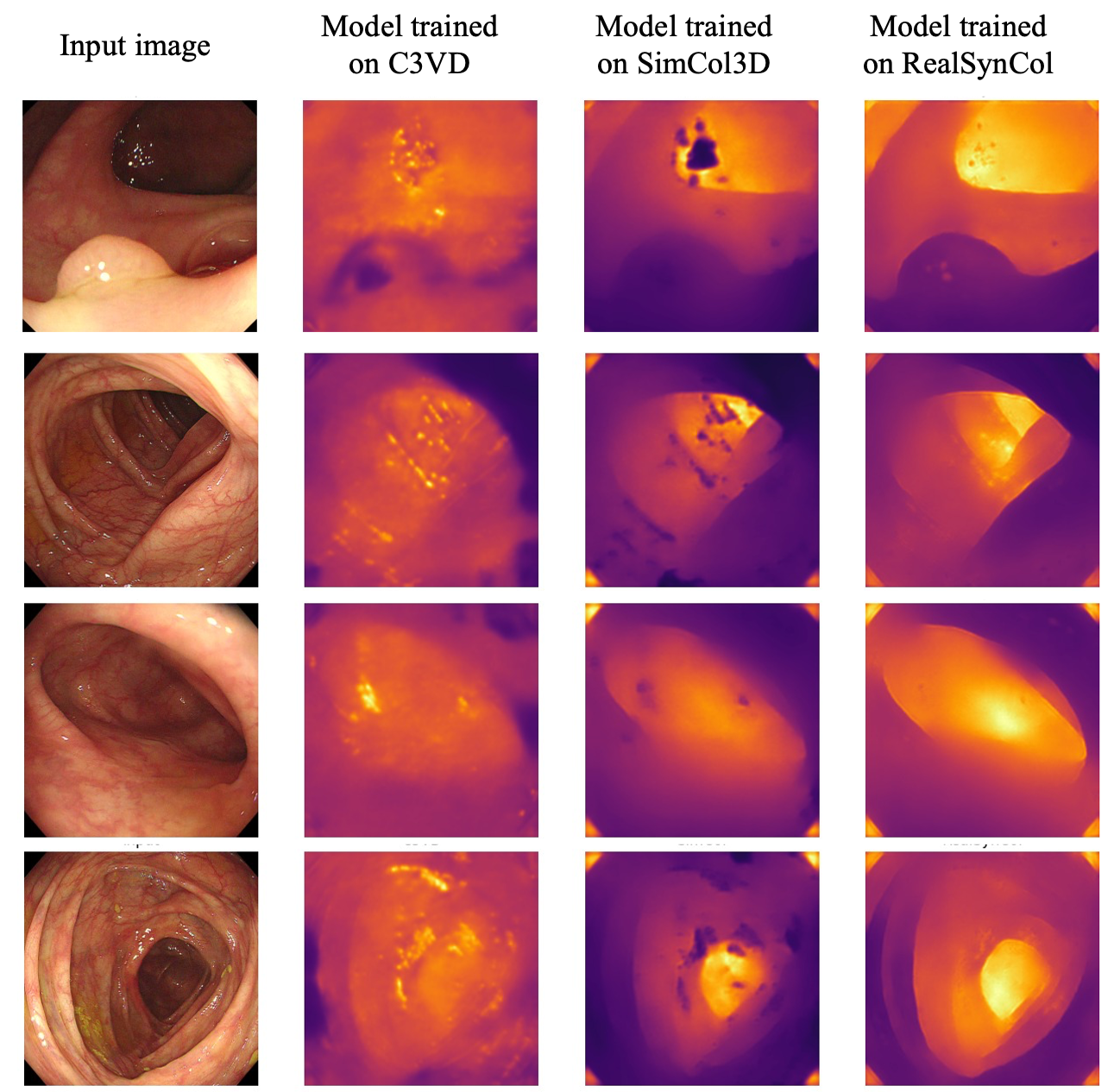}
\caption{Qualitative results of depth maps from clinical images predicted using the models trained on C3VD, SimCol3D, and RealSynCol.}
\label{fig:clinical_results}
\end{figure}

Acquiring ground-truth depth information from real colonoscopic images remains an open challenge. The colonic lumen presents a highly intricate geometry, with folds, sharp curvatures, and variable diameters, which hinder precise depth estimation. Complex illumination phenomena, such as specular reflections, shadows, and fluid artifacts, further compromise depth sensing, and conventional technologies for depth acquisition cannot be safely or effectively employed in vivo, leaving no feasible means to obtain dense and accurate depth maps during clinical procedures. As a result, the evaluation of depth estimation algorithms on clinical images must rely on qualitative assessment in the absence of ground-truth references. To further investigate the impact of training data on generalization and the ability to capture the characteristics of intraoperative images, depth prediction models were evaluated on clinical images from the SUN database~\cite{SUN}, and qualitative results are presented in Fig.~\ref{fig:clinical_results}.

These examples further confirm the tendency of the model trained on C3VD to underestimate depth, due to the very limited range of motion observed during training. In contrast, SimCol3D and RealSynCol can predict greater depth values with higher accuracy. However, the model trained on SimCol3D exhibits several artifacts around specular reflections, as these are not represented in the dataset and the model is therefore unable to compensate for them. On the other hand, the model trained on RealSynCol demonstrates high accuracy and robustness, with almost no artifacts around reflections or shadows, highlighting how a more realistic training dataset can significantly support the domain translation from virtual to real-world imagery.

\begin{figure}[!ht]
\centering 
\includegraphics[width=0.5\columnwidth]{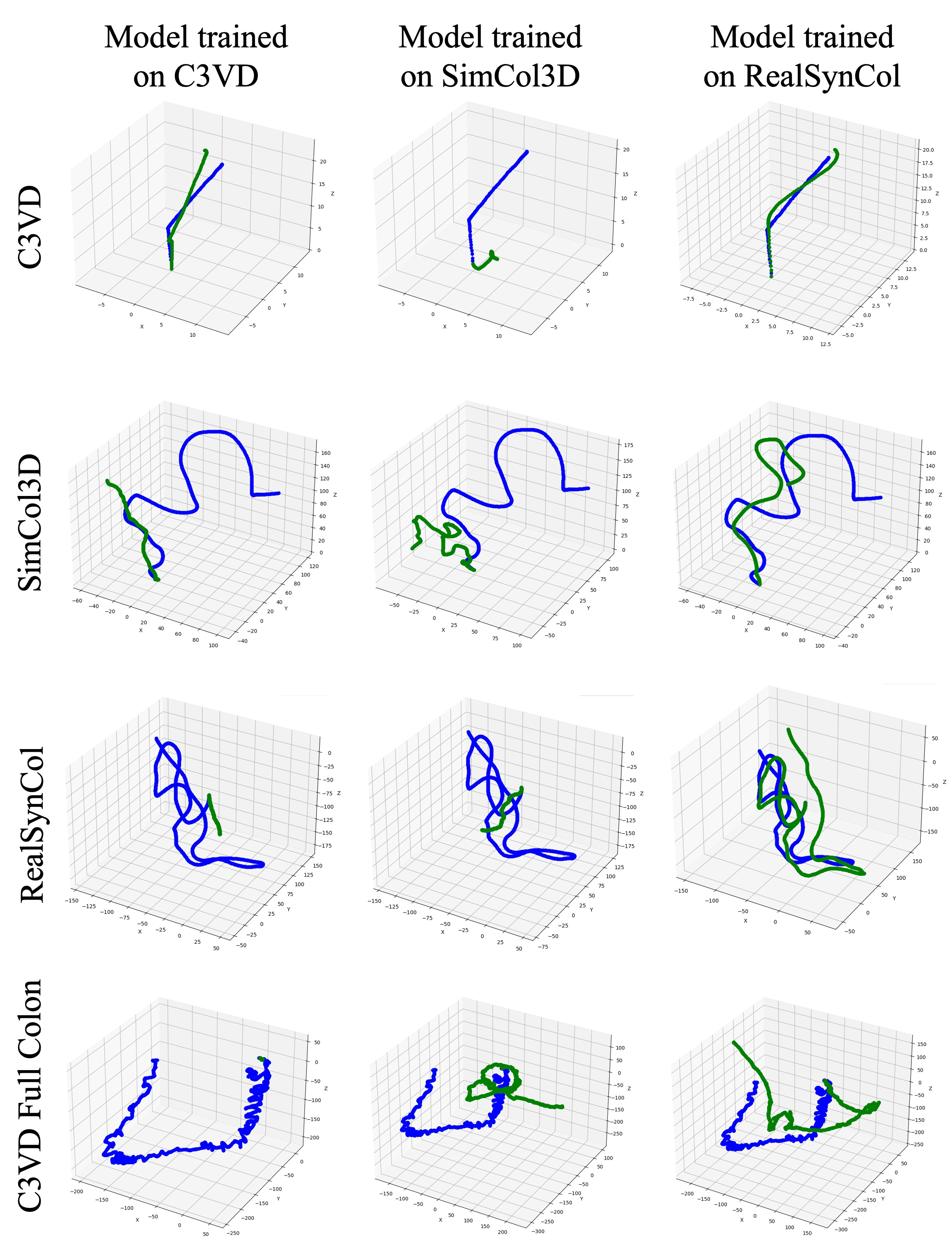}
\caption{Examples of poses predicted using the benchmark method. The first row represents sequence \textit{desc\_t4\_a} from C3VD, the second row depicts sequence \textit{O1} from SimCol3D, and the third row represents \textit{Colon\_9\_backward} from RealSynCol. The last row contains the trajectory of the entire colon \textit{seq1} from C3VD. The green represents the predicted trajectory, the blue represents the ground truth trajectory. The columns contain the predictions obtained using the model trained on C3VD, SimCol3D, and RealSynCol, respectively.}
\label{fig:pose_results}
\end{figure}

\begin{table*}[t]
\caption{Pose estimation results in terms of ATE, RTE, ROT, RMSE transl, RMSE rot. ATE, RTE, and ROT are reported as median values; in brackets, the interquartile range is given. For each sequence, the best metrics are in bold.}
\label{tab:pose_results}
\resizebox{\textwidth}{!}{
\begin{tabular}{lll>{\raggedleft}p{1.3cm} @{\hspace{0.1cm}} >{\raggedright}p{1.3cm}ccSc}
\textbf{Dataset} & \textbf{Model trained on} & \textbf{Sequence} & \multicolumn{2}{c}{\textbf{ATE (mm)} $\downarrow$} & \textbf{RTE (mm)} $\downarrow$ & \textbf{ROT (deg)} $\downarrow$ & \textbf{RMSE transl (mm)} $\downarrow$ & \textbf{RMSE rot (deg)} $\downarrow$ \\ \hline
C3VD & C3VD & desc\_t4\_a & 2.952 & (0.952) & \textbf{0.063} (\textbf{0.055}) & \textbf{0.070} (\textbf{0.069}) & 2.813 & \textbf{0.093} \\
C3VD & SimCol3D & desc\_t4\_a & 13.772 & (9.207) & 0.107 (0.144) & 0.253 (0.076) & 15.823 & 0.260 \\
C3VD & RealSynCol & desc\_t4\_a & \textbf{2.588} & \textbf{(1.181)} & 0.100 (0.081) & 0.487 (0.438) & \bfseries 2.659 & 0.554 \\ \hline
C3VD & C3VD & cecum\_t2\_a & \textbf{2.992} & \textbf{(1.024)} & \textbf{0.069} (\textbf{0.051}) & \textbf{0.054} (\textbf{0.055}) & \bfseries 6.665 & \textbf{0.070} \\
C3VD & SimCol3D & cecum\_t2\_a & 17.233 & (10.01) & 0.112 (0.103) & 0.106 (0.064) & 17.496 & 0.116 \\
C3VD & RealSynCol & cecum\_t2\_a & 17.106 & (14.384) & 0.113 (0.112) & 0.271 (0.148) & 18.846 & 0.328 \\ \hline
C3VD & C3VD & sigmoid\_t3\_a & 8.219 & (17.840) & \textbf{0.050} (\textbf{0.035}) & \textbf{0.055} (\textbf{0.050}) & 16.250 & \textbf{0.064} \\
C3VD & SimCol3D & sigmoid\_t3\_a & 14.057 & (15.562) & 0.080 (0.050) & 0.295 (0.111) & 18.254 & 0.320 \\
C3VD & RealSynCol & sigmoid\_t3\_a & \textbf{5.935} & \textbf{(5.720)} & 0.064 (0.037) & 0.189 (0.385) & \bfseries 9.422 & 0.344 \\ \hline
C3VD & C3VD & trans\_t3\_b & 4.065 & (1.211) & \textbf{0.037} (\textbf{0.028}) & \textbf{0.045} (\textbf{0.070}) & 4.295 & \textbf{0.056} \\
C3VD & SimCol3D & trans\_t3\_b & 3.896 & (0.666) & 0.040 (0.029) & 0.351 (0.065) & 3.862 & 0.349 \\
C3VD & RealSynCol & trans\_t3\_b & \textbf{2.930} & \textbf{(0.718)} & 0.048 (0.034) & 0.205 (0.118) & \bfseries 3.032 & 0.207 \\ \hline

C3VD & C3VD & seq\_1 & 213.919 & (152.966) & \textbf{0.384} (\textbf{0.489}) & 0.562(0.878) & 226.743 & 1.307 \\
C3VD & SimCol3D & seq\_1 & 197.227 & (118.090) & 0.503 (0.624) & \textbf{0.489} (\textbf{0.530}) & 204.180 & \textbf{1.066} \\
C3VD & RealSynCol & seq\_1 & \textbf{140.641} & \textbf{(90.018)} & 0.549 (0.639) & 0.600 (0.619) & \bfseries 161.752 & 1.100 \\ \hline

SimCol3D & C3VD & O1 & 59.126 & (95.133) & 0.608 (0.367) & 0.915 (1.137) & 91.351 & 1.462 \\
SimCol3D & SimCol3D & O1 & 79.603 & (128.406) & 0.794 (0.633) & 0.713 (0.673) & 110.814 & 0.999 \\
SimCol3D & RealSynCol & O1 & \textbf{29.338} & \textbf{(34.896)} & \textbf{0.459} (\textbf{0.260}) & \textbf{0.626} (\textbf{0.546}) & \bfseries 48.793 & \textbf{0.825} \\ \hline
SimCol3D & C3VD & O2 & 84.317 & (101.876) & 0.640 (0.349) & 0.827 (0.938) & 118.534 & 1.355 \\
SimCol3D & SimCol3D & O2 & 99.307 & (126.078) & 0.831 (0.600) & 0.654 (0.672) & 120.035 & 0.986 \\
SimCol3D & RealSynCol & O2 & \textbf{18.662} & \textbf{(16.318)} & \textbf{0.420} (\textbf{0.298}) & \textbf{0.588} (\textbf{0.525}) & \bfseries 22.603 & \textbf{0.810} \\ \hline
SimCol3D & C3VD & O3 & 81.639 & (113.173) & 0.693 (0.387) & 0.913 (1.251) & 107.196 & 1.479 \\
SimCol3D & SimCol3D & O3 & 73.991 & (104.433) & 0.666 (0.651) & \textbf{0.581} (\textbf{0.494}) & 102.594 & 0.935 \\
SimCol3D & RealSynCol & O3 & \textbf{37.530} & \textbf{(43.463)} & \textbf{0.511} (\textbf{0.271}) & 0.613 (0.521) & \bfseries 51.586 & \textbf{0.780} \\ \hline
RealSynCol & C3VD & Colon\_9\_backward & 139.674 & (47.959) & 1.062 (0.047) & 2.823 (2.370) & 136.929 & 3.749 \\
RealSynCol & SimCol3D & Colon\_9\_backward & 128.396 & (49.539) & 1.006 (0.037) & 1.278 (2.988) & 132.368 & 3.127 \\
RealSynCol & RealSynCol & Colon\_9\_backward & \textbf{20.532} & \textbf{(18.853)} & \textbf{0.179} (\textbf{0.151}) & \textbf{0.169} (\textbf{0.146}) & \bfseries 37.677 & \textbf{0.600} \\ \hline
RealSynCol & C3VD & Colon\_9\_forward & 124.384 & (103.734) & 1.030 (0.061) & 2.818 (2.516) & 139.617 & 3.734 \\
RealSynCol & SimCol3D & Colon\_9\_forward & 117.207 & (78.738) & 1.038 (0.256) & 1.849 (3.272) & 134.581 & 3.393 \\
RealSynCol & RealSynCol & Colon\_9\_forward & \textbf{33.567} & \textbf{(37.402)} & \textbf{0.231} (\textbf{0.153}) & \textbf{0.168} (\textbf{0.149}) & \bfseries 40.948 & \textbf{0.461}
\end{tabular}
}
\end{table*}

\subsubsection{Pose estimation}
Pose estimation results are detailed in Table~\ref{tab:depth_results}, whereas visual examples of the obtained trajectories are depicted in Fig.~\ref{fig:pose_results}. The metrics show significantly lower prediction errors for the model trained on RealSynCol, not only when evaluated on the same RealSynCol but also on SimCol3D, different sequences of C3VD, and the entire colon \textit{seq1} from the C3VD dataset. This clearly demonstrates that exposing the model to a broader and more realistic range of motions during training enables learning of more precise and accurate motion patterns, thereby enhancing its ability to generalize to previously unseen trajectories. The dataset yielding the best results, containing 28\,130 images, not only provides a larger amount of data but also encompasses a significantly wider diversity of camera trajectories. By contrast, the other datasets — SimCol3D (23\,421 frames) and C3VD (10\,015 frames) — include fewer samples and a narrower motion distribution, which jointly limit the model’s ability to capture the full variability of endoscopic motion and consequently reduce its generalization performance.

The sequence \textit{cecum\_t2\_a} from C3VD is the only instance where the model trained on C3VD outperforms the model trained on RealSynCol in terms of ATE. This can primarily be attributed to overfitting, as many sequences in C3VD depict trajectories from the cecum area. The analysis of all metrics on C3VD data reveals that the lowest values for relative translation and rotation errors are often achieved by the C3VD-trained model. However, as illustrated in Fig.~\ref{fig:pose_results}, this does not necessarily indicate that the relative translations or rotations are more accurate. The model trained on C3VD has been exposed to limited movements and rotations, which leads to a tendency to significantly underestimate the extent of motion, predicting near-static transformations between consecutive frames. Mathematically, this can result in lower error values, even when the predictions are incorrect.

Trajectories in SimCol3D are typically simpler and smoother compared to those in RealSynCol. Self-supervised methods like Lite-Mono require a broader diversity of rotations and motions to effectively learn endoscopic trajectory prediction. For pose estimation, the findings suggest that motion patterns in SimCol3D and C3VD are overly simplified, constraining the model’s learning capacity. This is further clarified by the test performed on the entire trajectory of \textit{seq1} from the C3VD dataset. 
The model trained on C3VD, despite learning from visually similar images, fails to capture the full range of motion, resulting in a collapsed trajectory with closely clustered points. The model trained on SimCol3D struggles to predict movements more complex than those seen during training. In contrast, the model trained on RealSynCol, though affected by rotational inaccuracies likely due to the noisy endoscope motion, successfully captures a wider and more extended trajectory, closely approximating the ground truth shape. 
A statistical analysis was conducted to evaluate the significance of the pose estimation results. For each sequence, the ATE values obtained with the models trained on the three datasets were compared using the non-parametric Wilcoxon signed-rank test with Bonferroni correction. The analysis reveals significant differences among the three methods across all sequences, with the exception of \textit{desc\_t4\_a} in the C3VD dataset. As shown in the first row of Fig. \ref{fig:pose_results}, the results obtained with the model trained on C3VD and RealSynCol are very similar in this sequence, and the statistical analysis indicates no significant difference. This further highlights the generalization capabilities conferred by the RealSynCol training dataset, yielding results comparable to those obtained with the C3VD dataset.

\section{Data Availability}
The dataset is available at the following link: RealSynCol Dataset link, 
 together with the scripts for reading the files and converting the poses from Blender to OpenCV reference frame, and the texture pattern used for dataset generation. 

\section{Code Availability}
We conducted all benchmark and ablation experiments using the original Lite-Mono codebase and default settings, adapting only the dataloader to match the directory structures of the considered datasets. The code used to fine-tune the original DAM v2, the Lite-Mono adaptation, and further implementation details are available at: \url{https://github.com/ChiaraLena/RealSynCol}.

\section{Funding}
This work was supported by the Multilayered
Urban Sustainability Action (MUSA) project (ECS00000037), funded by the European Union – NextGenerationEU, under the National Recovery and Resilience Plan (NRRP), the ANTHEM project, funded by the National Plan for NRRP
Complementary Investments (CUP: B53C22006700001), and in part by the National Institute of Biomedical Imaging and Bioengineering (NIBIB), USA of the National Institutes of Health (NIH) under award number 2R01EB018992 and by Cancer Research UK (CRUK) Early Detection and Diagnosis Research Committee under grant number 27744.  Any opinions, findings, conclusions, or recommendations expressed in this publication are those of the authors and do not necessarily reflect the views of the NIH or CRUK.

\bibliographystyle{unsrt}
\bibliography{main}

\end{document}